\documentclass[10pt,twocolumn,letterpaper]{article}

\usepackage[bottom]{footmisc}

\usepackage{enumitem}
\usepackage[dvipsnames]{xcolor}

\usepackage{tabularx}

\usepackage{cvpr}
\usepackage{times}
\usepackage{epsfig}
\usepackage{graphicx}
\usepackage{amsmath}
\usepackage{amssymb}
\usepackage{outlines}
\usepackage{siunitx}

\newcommand*\colourcheck[1]{%
  \expandafter\newcommand\csname #1check\endcsname{\textcolor{#1}{\ding{52}}}%
}
\newcommand*\colourcross[1]{%
  \expandafter\newcommand\csname #1cross\endcsname{\textcolor{#1}{\ding{56}}}%
}

\colourcheck{green}
\colourcheck{yellow}
\colourcross{red}

\usepackage{tabularx} 
\newcolumntype{L}{>{\raggedright\arraybackslash}X}
\newcolumntype{R}{>{\raggedleft\arraybackslash}X}
\newcolumntype{C}{>{\centering\arraybackslash}X} 

\usepackage{color, colortbl} 
\definecolor{LightGray}{rgb}{0.9,0.9,0.9}

\usepackage{multirow}

\newcommand\Tstrut{\rule{0pt}{2.6ex}}         

\usepackage{graphicx} 
\usepackage{animate}

\usepackage{pifont}

\usepackage[skip=1pt]{caption} 
\usepackage{subcaption}
\usepackage{color}

\usepackage[pagebackref=true,breaklinks=true,letterpaper=true,colorlinks,bookmarks=false]{hyperref}

\cvprfinalcopy 

\def\cvprPaperID{3301} 

\definecolor{somegray}{rgb}{0.5, 0.5, 0.5}
\newcommand{\darkgrayed}[1]{\textcolor{somegray}{#1}}
\makeatletter
\newcommand*\titleheader[1]{\gdef\@titleheader{#1}}
\AtBeginDocument{%
  \let\st@red@title\@title
  \def\@title{%
    \vskip-3em
    \bgroup\normalfont\large\centering\@titleheader\par\egroup
    \vskip1.5em\st@red@title}
}
\makeatother

\titleheader{\darkgrayed{This paper has been accepted for publication at the\\
IEEE Conference on Computer Vision and Pattern Recognition (CVPR), Nashville, 2021.
\copyright IEEE}}

\makeatletter
\let\@oldmaketitle\@maketitle
\renewcommand{\@maketitle}{\@oldmaketitle

\centering
\includegraphics[width=1.0\textwidth]{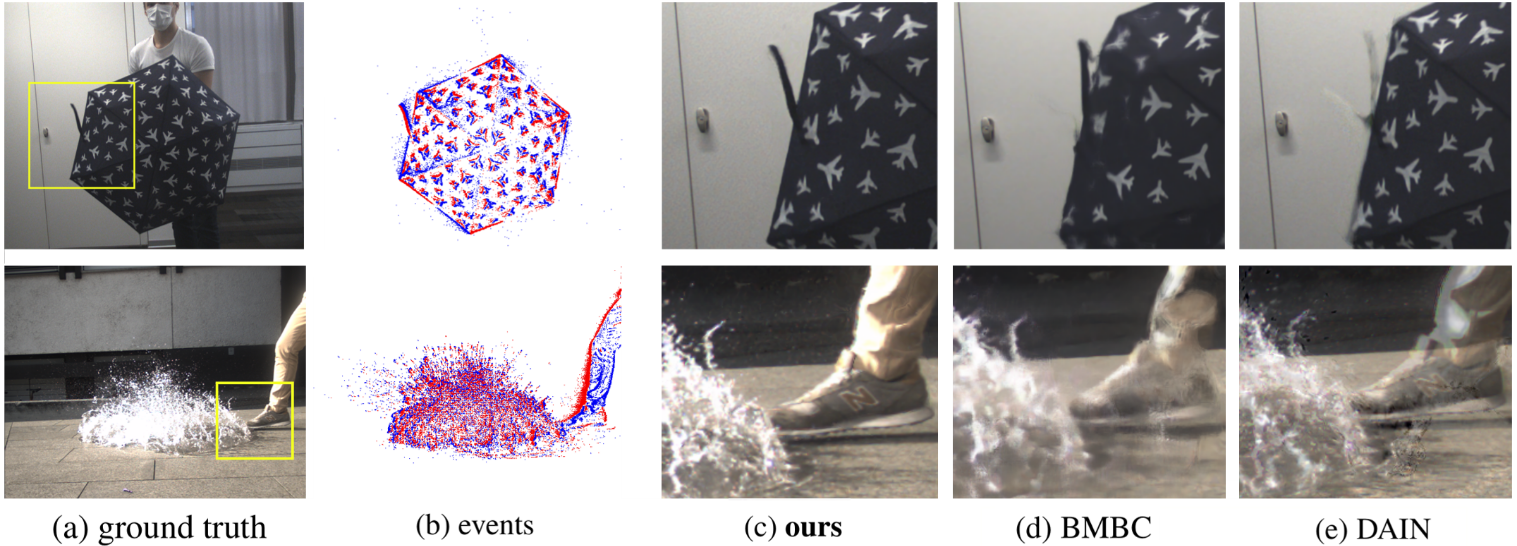}
\vspace{1pt}
\captionof{figure}{Qualitative results comparing our proposed method, Time Lens, with DAIN~\cite{bao_2019} and BMBC~\cite{park_2020}. Our method can interpolate frames in highly-dynamic scenes, such as while spinning an umbrella (top row) and bursting a balloon (bottom row). It does this by combining events (b) and frames (a).    
\vspace{0.3cm}
}\label{fig:eyecatcher}
}
\makeatother
\title{Time Lens: Event-based Video Frame Interpolation}

\begin{document}
\newcommand{\rev}[1]{#1}
\newcommand{\reviewStepan}[1]{#1}

\author{Stepan Tulyakov\thanks{\reviewStepan{indicates equal contribution}}$^{*,1}$ \quad Daniel Gehrig$^{*,2}$ \quad Stamatios Georgoulis$^1$ \quad Julius Erbach$^1$ \and Mathias Gehrig$^2$ \quad Yuanyou Li$^1$ \quad Davide Scaramuzza$^2$\\
$^1$Huawei Technologies, Zurich Research Center\\
$^2$Dept. of Informatics, Univ. of Zurich and Dept. of Neuroinformatics, Univ. of Zurich and ETH Zurich
\vspace{-3ex}
}

\maketitle
\thispagestyle{empty}
\pagestyle{empty}

\begin{abstract}
\vspace{-3pt}
State-of-the-art frame interpolation methods generate intermediate frames by inferring object motions in the image from consecutive key-frames. In the absence of additional information, first-order approximations, i.e. optical flow, must be used, but this choice restricts the types of motions that can be modeled, leading to errors in highly dynamic scenarios.
Event cameras are novel sensors that address this limitation by providing auxiliary visual information in the blind-time between frames.
They asynchronously measure per-pixel brightness changes and do this with high temporal resolution and low latency.
Event-based frame interpolation methods typically adopt a synthesis-based approach, where predicted frame residuals are directly applied to the key-frames. However, while these approaches can capture non-linear motions they suffer from ghosting and perform poorly in low-texture regions with few events. 
Thus, synthesis-based and flow-based approaches are complementary. 
In this work, we introduce \emph{Time Lens}, a novel method that leverages the advantages of both.
We extensively evaluate our method on three synthetic and two real benchmarks where we show an up to 5.21 dB improvement in terms of PSNR over state-of-the-art frame-based and event-based methods. 
Finally, we release a new large-scale dataset in highly dynamic scenarios, aimed at pushing the limits of existing methods. \\

\vspace{-2ex}

\end{abstract}
\section*{Multimedia Material} 
The High-Speed Event and RGB~(HS-ERGB) dataset and evaluation code can be found at: \url{http://rpg.ifi.uzh.ch/timelens}
\section{Introduction}
Many things in real life can happen in the blink of an eye. A hummingbird flapping its wings, a cheetah accelerating towards its prey, a tricky stunt with the skateboard, or even a baby taking its first steps. Capturing these moments as high-resolution videos with high frame rates typically requires professional high-speed cameras, that are inaccessible to casual users. Modern mobile device producers have tried to incorporate more affordable sensors with similar functionalities into their systems, but they still suffer from the large memory requirements and high power consumption associated with these sensors. 

Video Frame Interpolation~(VFI) addresses this problem, by converting videos with moderate frame rates high frame rate videos in post-processing. In theory, any number of new frames can be generated between two keyframes of the input video. Therefore, VFI is an important problem in video processing with many applications, ranging from super slow motion~\cite{jiang_2018} to video compression~\cite{wu_2018}. 

\textbf{Frame-based interpolation} approaches relying solely on input from a conventional frame-based camera that records frames synchronously and at a fixed rate. There are  several classes of such methods that we describe below. 

\textit{Warping-based approaches}~\cite{niklaus_2018,jiang_2018,xue_2019,niklaus_2020,park_2020} combine optical flow estimation~\cite{ilg_2017,liu_2017,sun_2018} with image warping~\cite{jaderberg_2015}, to generate intermediate frames in-between two consecutive key frames. 
More specifically, under the assumptions of linear motion and brightness constancy between frames, these works compute optical flow and warp the input keyframe(s) to the target frame, while leveraging concepts, like contextual information~\cite{niklaus_2018}, visibility maps~\cite{jiang_2018}, spatial transformer networks~\cite{xue_2019}, forward warping~\cite{niklaus_2020}, or dynamic blending filters~\cite{park_2020}, to improve the results. While most of these approaches assume linear motion, some recent works assume quadratic~\cite{xu_2019} or cubic~\cite{Chi20all} motions. Although these methods can address non-linear motions, they are still limited by their order, failing to capture arbitrary motion. 


\textit{Kernel-based approaches}~\cite{niklaus_2017a,niklaus_2017b} avoid the explicit motion estimation and warping stages of warping-based approaches. Instead, they model VFI as local convolution over the input keyframes, where the convolutional kernel is estimated from the keyframes. This approach is more robust to motion blur and light changes. Alternatively, \textit{phase-based approaches}~\cite{meyer_2018} pose VFI as a phase shift estimation problem, where a neural network decoder directly estimates the phase decomposition of the intermediate frame. However, while these methods can in theory model arbitrary motion, in practice they do not scale to large motions due to the locality of the convolution kernels.

In general, all frame-based approaches assume simplistic motion models (e.g. linear) due to the absence of visual information in the blind-time between frames, which poses a fundamental limitation of purely frame-based VFI approaches. In particular, the simplifying assumptions rely on brightness and appearance constancy between frames, which limits their applicability in highly dynamic scenarios such as (\emph{i}) for non-linear motions between the input keyframes, (\emph{ii}) when there are changes in illumination or motion blur, and (\emph{iii}) non-rigid motions and new objects appearing in the scene between keyframes.

\textbf{Multi-camera approaches.} 
To overcome this limitation, some works seek to combine inputs from several frame-based cameras with different spatio-temporal trade-offs. For example, \cite{gupta_2009} combined low-resolution video with high resolution still images, whereas~\cite{paliwal_2020} fused a low-resolution high frame rate video with a high resolution low frame rate video. Both approaches can recover the missing visual information necessary to reconstruct true object motions, but this comes at the cost of a bulkier form factor, higher power consumption, and a larger memory footprint.
%

\textbf{Event-based approaches.} Compared to standard frame-based cameras, event cameras~\cite{Lichtsteiner08ssc,brandli_2014} do not incur the aforementioned costs. 
They are novel sensors that only report the per-pixel intensity changes, as opposed to the full intensity images and do this with high temporal resolution and low latency on the order of microseconds. The resulting output is an asynchronous stream of binary ``events" which can be considered a compressed representation of the true visual signal. 
These properties render them useful for VFI under highly dynamic scenarios (e.g. high-speed non-linear motion, or challenging illumination).  

\textit{Events-only approaches} reconstruct high frame rate videos directly from the stream of incoming events using GANs~\cite{wang_2019a}, RNNs~\cite{rebecq_2019a,rebecq_2019b,scheerlinck_2020}, or even self-supervised CNNs~\cite{paredes_2020}, and can be thought of as a proxy to the VFI task. However, since the integration of intensity gradients into an intensity frame is an ill-posed problem, the global contrast of the interpolated frames is usually miscalculated. Moreover, as in event cameras intensity edges are only exposed when they move, the interpolation results are also dependent on the motion.


\textit{Events-plus-frames approaches.} 
As certain event cameras such as the Dynamic and Active VIsion Sensor (DAVIS)~\cite{brandli_2014} can simultaneously output the event stream and intensity images -- the latter at low frame rates and prone to the same issues as frame-based cameras (e.g. motion blur) -- several works~\cite{pan_2019,wang_2019b,jiang_2020,wang_2020} use both streams of information. Typically, these works tackle VFI in conjunction with de-blurring, de-noising, super-resolution, or other relevant tasks. They synthesize intermediate frames by accumulating temporal brightness changes, represented by events, from the input keyframes and applying them to the key frames. While these methods can handle illumination changes and non-linear motion they still perform poorly compared to the frame-based methods~(please see \S~\ref{sec:benchmarking}), as due to the inherent instability of the contrast threshold and sensor noise, not all brightness changes are accurately registered as events.       

\textbf{Our contributions} are as follows 

\begin{enumerate}[noitemsep,topsep=0pt,parsep=0pt,partopsep=0pt]
    \item We address the limitations of all aforementioned methods by introducing a CNN framework, named \textit{Time Lens}, that marries the advantages of warping- and synthesis-based interpolation approaches. In our framework, we use a synthesis-based approach to ground and refine results of high-quality warping-based approach and provide the ability to handle illumination changes and new objects appearing between keyframes~(refer Fig.~\ref{fig:qualitative}), 
    
    \item \reviewStepan{We introduce a new warping-based interpolation approach that estimates motion from events, rather than frames and thus has several advantages: it is more robust to motion blur and can estimate non-linear motion between frames. Moreover, the proposed method provides a higher quality interpolation compared to synthesis-based methods that use events when event information is not sufficient or noisy.}

    \item We empirically show that the proposed Time Lens greatly outperforms state-of-the-art frame-based and event-based methods, published over recent months, on three synthetic and two real benchmarks where we show an up to 5.21 dB improvement in terms of PSNR. 

\end{enumerate}

\section{Method}
\begin{figure}[t]
    \centering
	\includegraphics[width=0.45\textwidth]{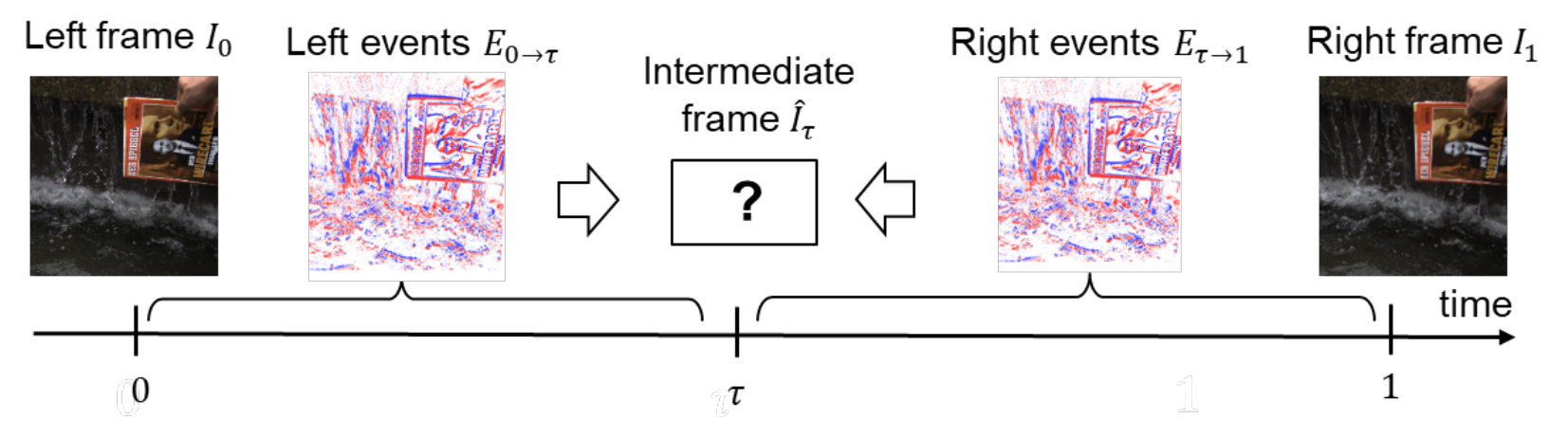}
	\caption{Proposed event-based VFI approach.}
	\label{fig:task}
\end{figure}

\textbf{Problem formulation.} Let us assume an event-based VFI setting, where we are given as input the left $I_0$ and right $I_1$ RGB key frames, as well as the left $E_{0\rightarrow \tau}$ and right $E_{\tau \rightarrow 1}$ \textit{event sequences}, and we aim to interpolate (one or more) new frames $\hat{I_{\tau}}$ at random timesteps $\tau$ in-between the key frames. Note that, the event sequences ($E_{0\rightarrow \tau}$, $E_{\tau \rightarrow 1}$) contain all asynchronous events that are triggered from the moment the respective (left $I_0$ or right $I_1$) key RGB frame is synchronously sampled, till the timestep $\tau$ at which we want to interpolate a new frame $\hat{I_{\tau}}$. Fig.~\ref{fig:task} illustrates the proposed event-based VFI setting.

\begin{figure*}[t]
    \centering
    \begin{subfigure}[a]{0.56\textwidth}
	    \centering
	    \includegraphics[width=\textwidth]{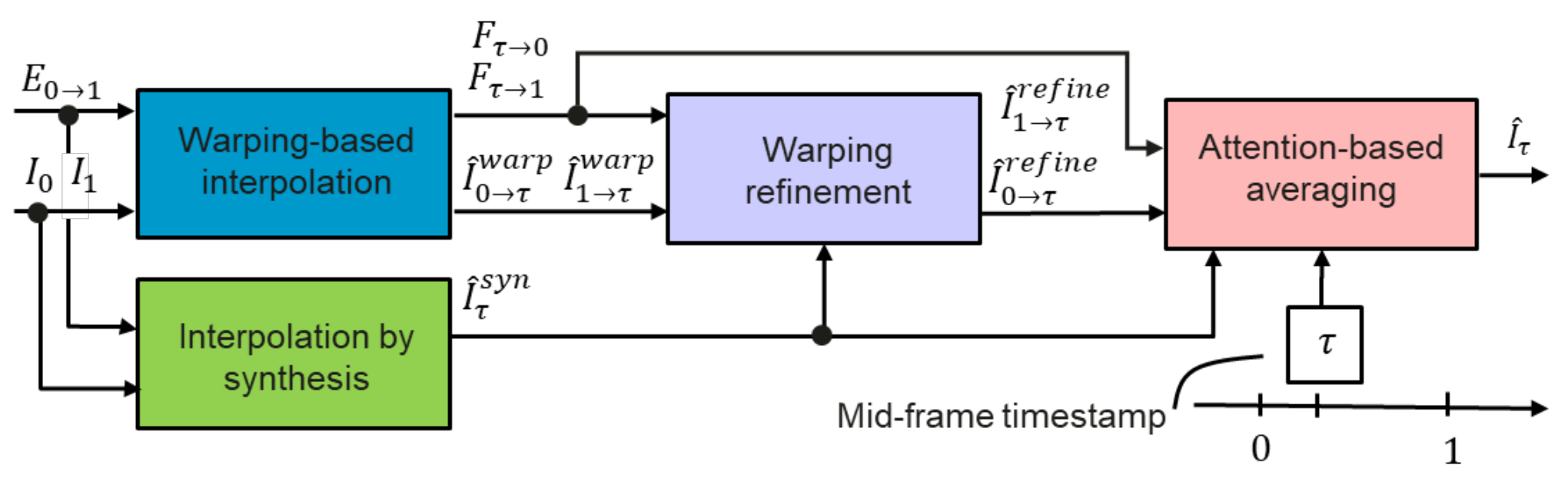}
	    \caption{Overview of the proposed method.}
	    \label{fig:workflow}
    \end{subfigure}
    \hfill    \begin{subfigure}[a]{0.43\textwidth}
	    \centering
	    \includegraphics[width=\textwidth]{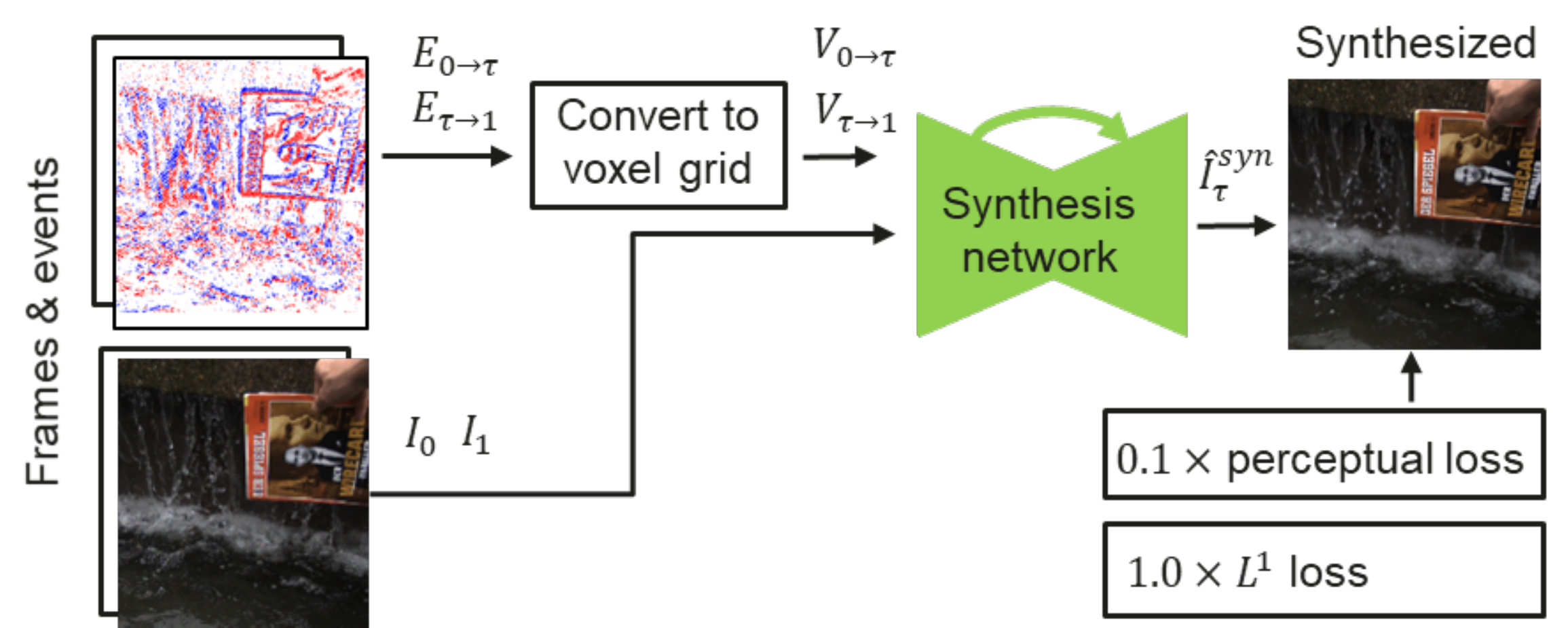}
	    \caption{Interpolation by synthesis module.}
	    \label{fig:synthesis}
    \end{subfigure}
        \begin{subfigure}[a]{0.56\textwidth}
	    \centering
	    \includegraphics[width=\textwidth]{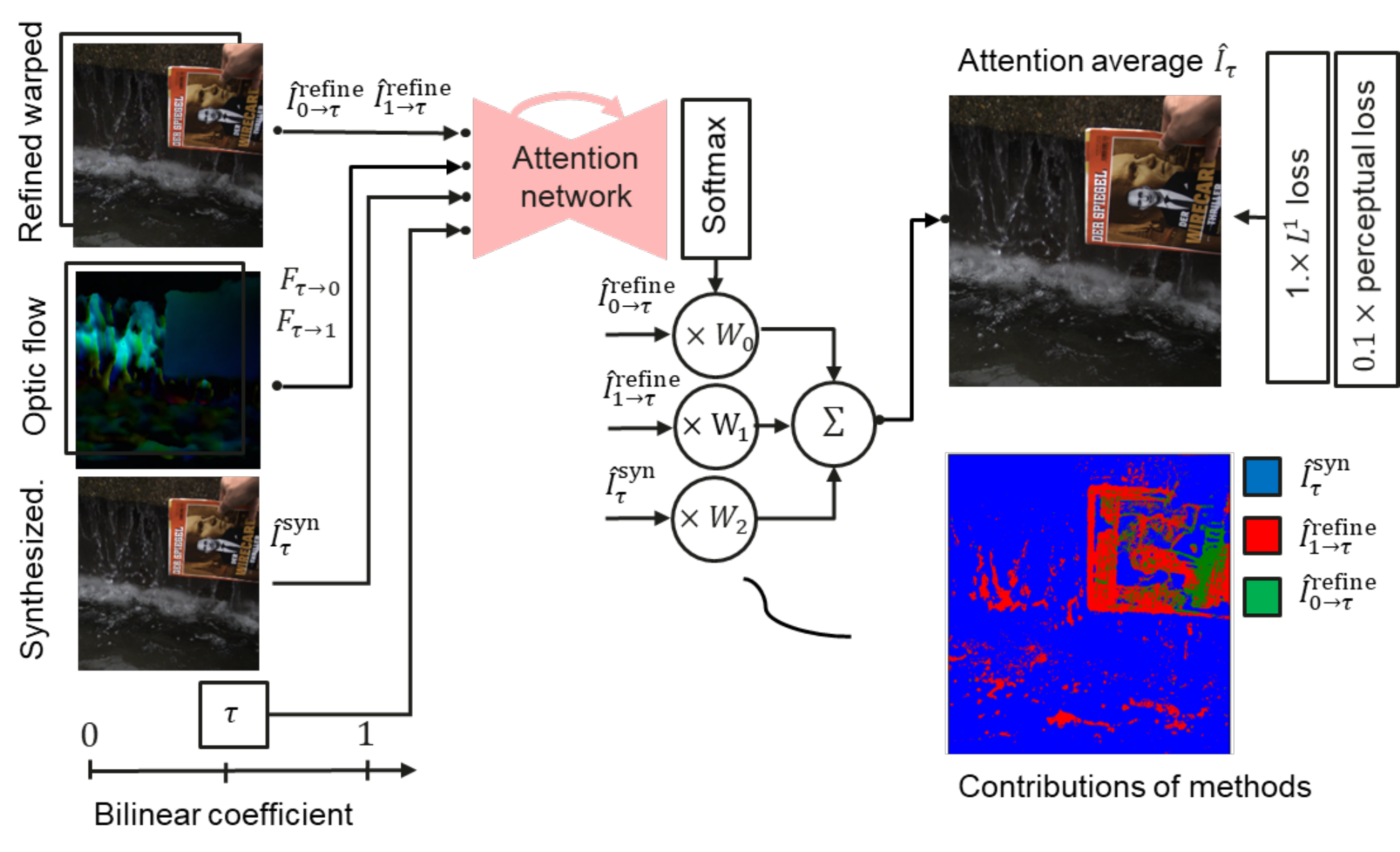}
	    \caption{Attention-based averaging module.}
	    \label{fig:attention}
    \end{subfigure}
    \hfill
    \begin{subfigure}[a]{0.43\textwidth}
	    \centering
	    \includegraphics[width=\textwidth]{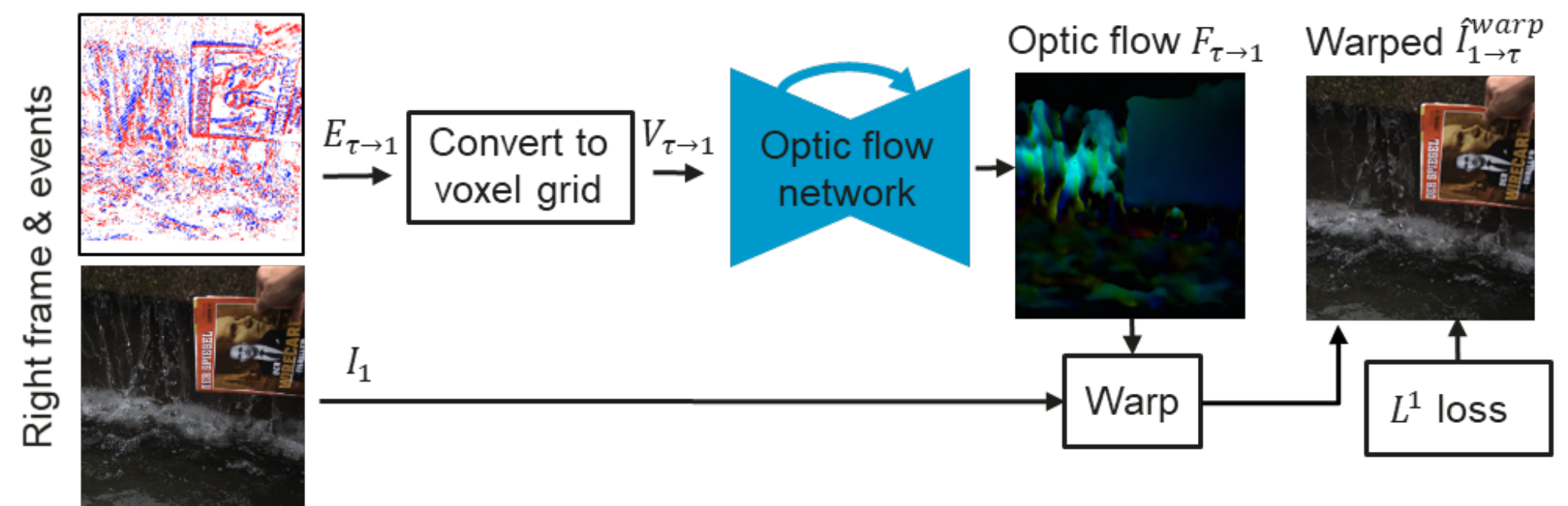}
	    \caption{Warping-based interpolation module.}
	    \label{fig:warp}
	    \includegraphics[width=\textwidth]{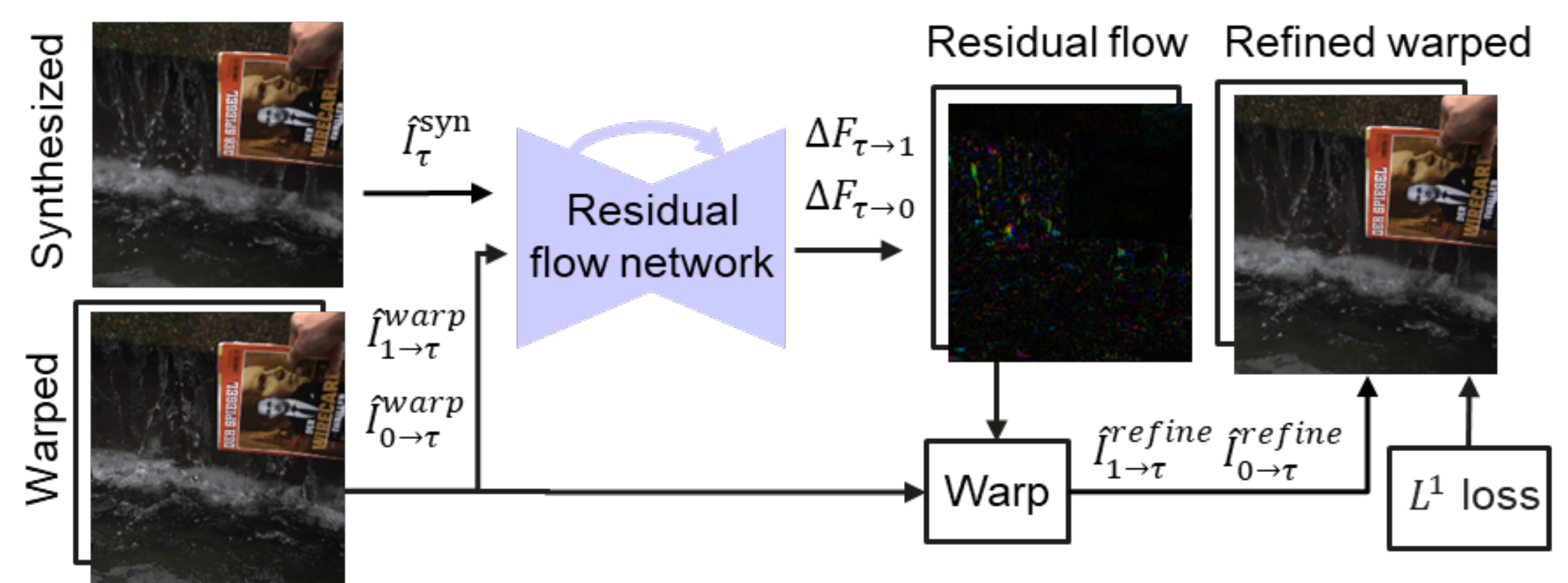}
	    \caption{Warping refinement module.}
	    \label{fig:refinement}
    \end{subfigure}
    \caption{Structure of the proposed method. The overall workflow of the method is shown in Fig.~\ref{fig:workflow} and individual modules are shown in Fig.~\ref{fig:warp},~\ref{fig:synthesis},~\ref{fig:refinement} and~\ref{fig:attention}. In the figures we also show loss function that we use to train each module. We show similar modules in the same color across the figures.}
\end{figure*}

\textbf{System overview.} To tackle the problem under consideration we propose a learning-based framework, namely \textit{Time Lens}, that consists of four dedicated modules that serve complementary interpolation schemes, i.e. warping-based and synthesis-based interpolation. In particular, (1) the \textit{warping-based interpolation} module estimates a new frame by warping the boundary RGB keyframes using optical flow estimated from the respective event sequence; (2) the \textit{warping refinement} module aims to improve this estimate by computing residual flow; (3) the \textit{interpolation by synthesis} module estimates a new frame by directly fusing the input information from the boundary keyframes and the event sequences; finally (4) the \textit{attention-based averaging} module aims to optimally combine the warping-based and synthesis-based results. In doing so, Time Lens marries the advantages of warping- and synthesis-based interpolation techniques, allowing us to generate new frames with color and high textural details while handling non-linear motion, light changes, and motion blur. The workflow of our method is shown in Fig.~\ref{fig:workflow}.

All modules of the proposed method use the same backbone architecture, which is an hourglass network with skip connections between the contracting and expanding parts, similar to~\cite{jiang_2018}. The backbone architecture is described in more detail in the supplementary materials. Regarding the learning representation~\cite{Gehrig19iccv} used to encode the event sequences, all modules use the \textit{voxel grid} representation. Specifically, for event sequence $E_{\tau_0 \rightarrow \tau_{end}}$ we compute a voxel grid $V_{\tau_0 \rightarrow \tau_{end}}$ following the procedure described in~\cite{zihao_2018}. In the following paragraphs, we analyze each module and its scope within the overall framework.

\textbf{Interpolation by synthesis}, as shown in Fig.~\ref{fig:synthesis}, directly regresses a new frame $\hat{I}^{\text{syn}}$ given the left $I_0$ and right $I_1$ RGB keyframes and events sequences $E_{0\rightarrow \tau}$ and $E_{\tau \rightarrow 1}$ respectively. The merits of this interpolation scheme lie in its ability to handle changes in lighting\reviewStepan{, such as water reflections in Fig.~\ref{fig:rebuttal}} and a sudden appearance of new objects in the scene, because unlike warping-based method, it does not rely on the brightness constancy assumption. Its main drawback is the distortion of image edges and textures when event information is noisy or insufficient \reviewStepan{because of high contrast thresholds, e.g. triggered by the book in Fig.~\ref{fig:rebuttal}}.

\begin{figure}[tbh]
    \centering
	\includegraphics[width=0.45\textwidth]{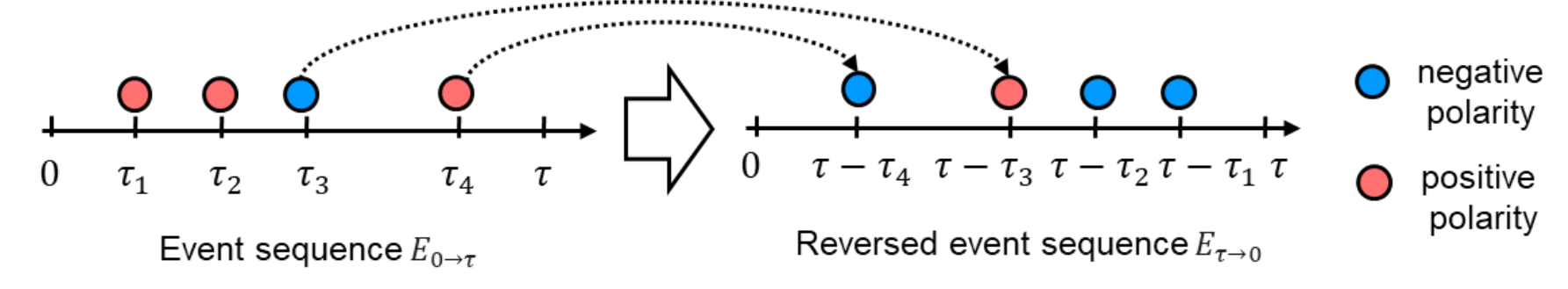}
	\caption{Example of an event sequence reversal.}
	\label{fig:reversal}
\end{figure}

\textbf{Warping-based interpolation}, \reviewStepan{shown in Fig.~\ref{fig:warp}}, first estimates the optical flow $F_{\tau \rightarrow 0}$ and $F_{\tau \rightarrow 1}$ between a latent new frame $\hat{I_{\tau}}$ and boundary keyframes $I_0$ and $I_1$ \reviewStepan{using events $E_{\tau\rightarrow 0}$ and $E_{\tau\rightarrow 1}$ respectively. We compute $E_{\tau\rightarrow 0}$, by reversing the event sequence $E_{0 \rightarrow \tau}$, as shown in Fig.~\ref{fig:reversal}.} Then our method uses computed optical flow to warp the boundary keyframes in timestep $\tau$ using differentiable interpolation~\cite{jaderberg_2015}, which in turn produces two new frame estimates $\hat{I}^{\text{warp}}_{0\rightarrow \tau}$ and $\hat{I}^{\text{warp}}_{1\rightarrow \tau}$.

\reviewStepan{The major difference of our approach from the traditional warping-based interpolation methods~\cite{niklaus_2018,jiang_2018,niklaus_2020, xu_2019}, is that the latter compute optical flow between keyframes using the frames themselves and then approximate optical flow between the latent middle frame and boundary by using a linear motion assumption. This approach does not work when motion between frames is non-linear and keyframes suffer from motion blur. By contrast, our approach computes the optical flow from the events, and thus can naturally handle blur and non-linear motion. Although events are sparse, the resulting flow is sufficiently
dense as shown in Fig.~\ref{fig:warp}, especially in textured areas with dominant mostion, which is most important for interpolation.}

\reviewStepan{Moreover, the warping-based interpolation approach relying on events also works better than synthesis-based method in the scenarios when event data is noisy or not sufficient due to high contrast thresholds, e.g. the book in Fig.~\ref{fig:rebuttal}. On the down side, this method still relies on the brightness constancy assumption for optical flow estimation and thus can not handle brightness changes and new objects appearing between keyframes, e.g. water reflections in Fig.~\ref{fig:rebuttal}.}

\textbf{Warping refinement} module computes refined interpolated frames, $\hat{I}^{\text{refine}}_{0\rightarrow \tau}$ and $\hat{I}^{\text{refine}}_{1\rightarrow \tau}$, by estimating residual optical flow, $\Delta F_{\tau\rightarrow 0}$ and $\Delta F_{\tau\rightarrow 1}$ respectively, between the warping-based interpolation results, $\hat{I}^{\text{warp}}_{0\rightarrow \tau}$ and $\hat{I}^{\text{warp}}_{1\rightarrow \tau}$, and the synthesis result $\hat{I}^{\text{syn}}_{\tau}$. It then proceeds by warping $\hat{I}^{\text{warp}}_{0\rightarrow \tau}$ and $\hat{I}^{\text{warp}}_{1\rightarrow \tau}$ for a second time using the estimated residual optical flow, as shown in Fig.~\ref{fig:refinement}. The refinement module draws inspiration from the success of optical flow and disparity refinement modules in~\cite{ilg_2017,pang_2017}, and also by our observation that the synthesis interpolation results are usually perfectly aligned with the ground-truth new frame. Besides computing residual flow, the warping refinement module also performs inpainting of the occluded areas, by filling them with values from nearby regions.   

Finally, the \textbf{attention averaging} module, \reviewStepan{shown in Fig.~\ref{fig:attention}}, blends in a pixel-wise manner the results of synthesis $\hat{I}^{\text{syn}}_{\tau}$ and warping-based interpolation $\hat{I}^{\text{refine}}_{0\rightarrow \tau}$ and $\hat{I}^{\text{refine}}_{1\rightarrow \tau}$ to achieve final interpolation result $\hat{I}_{\tau}$. \reviewStepan{This module leverages the complementarity of the warping- and synthesis-based interpolation methods and produces a final result, which is better than the results of both
methods by 1.73 dB in PSNR as shown in Tab.~\ref{tab:ablation} and illustrated in Fig.~\ref{fig:rebuttal}.} 

A similar strategy was used in~\cite{niklaus_2020,jiang_2018}, however these works only blended the warping-based interpolation results to fill the occluded regions, while we blend both warping and synthesis-based results, and thus can also handle light changes. We estimate the blending coefficients using an attention network that takes as an input the interpolation results, $\hat{I}^{\text{refine}}_{0 \rightarrow \tau}$, $\hat{I}^{\text{refine}}_{1 \rightarrow \tau}$ and $\hat{I}^{\text{syn}}$, the optical flow results $F_{\tau \rightarrow 0}$ and $F_{ \tau \rightarrow 1}$ and bi-linear coefficient $\tau$, that depends on the position of the new frame as a channel with constant value.  

\subsection{High Speed Events-RGB~(HS-ERGB) dataset}\label{sec:high_speed_events_and_rgb}
\begin{figure}[t]
	\centering
	\includegraphics[width=0.7\columnwidth]{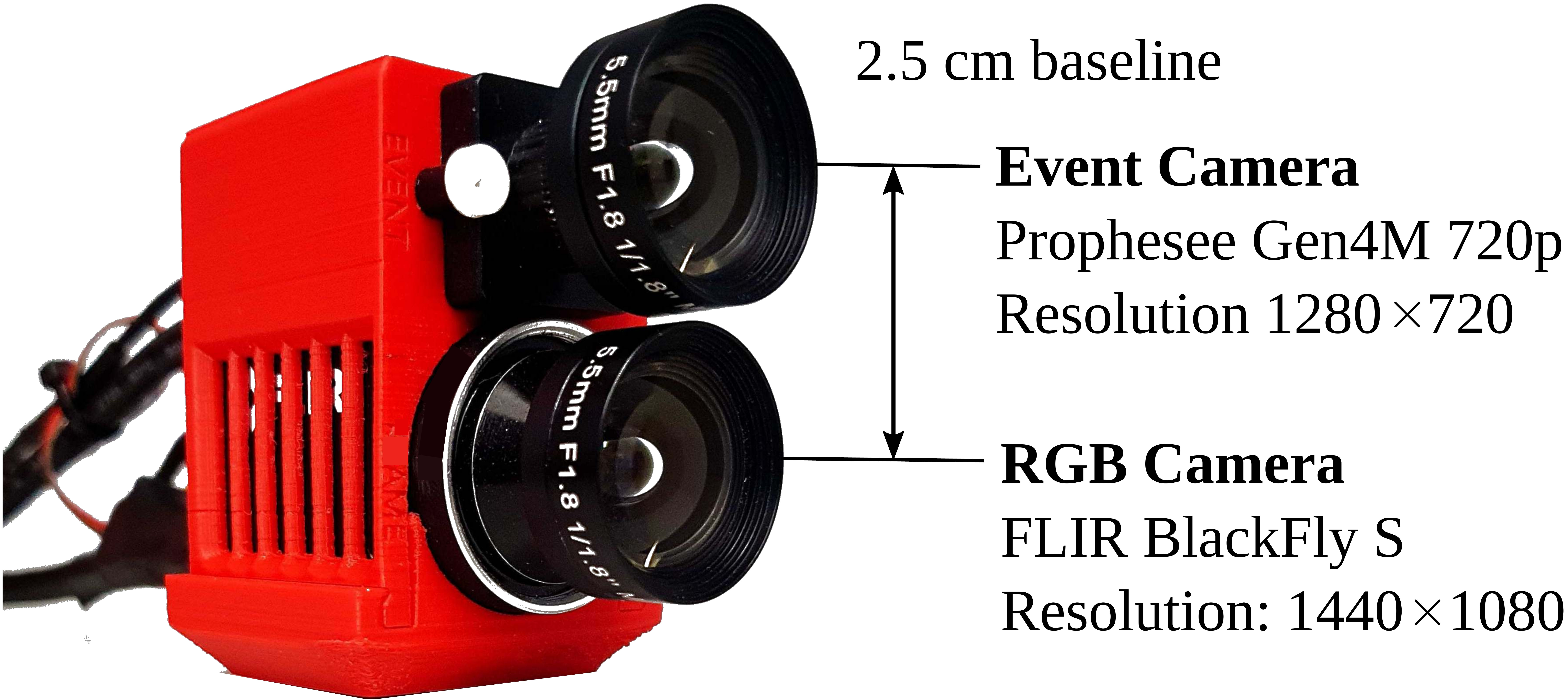}
    \caption{Illustration of the dual camera setup. 
    It comprises a Prophesee Gen4 720p monochrome event camera (top) and a FLIR BlackFly S RGB camera (bottom). 
    Both cameras are hardware synchronized with a baseline of \SI{2.5}{\centi\meter}}.
	\label{fig:dual_setup}
\end{figure}

Due to the lack of available datasets that combine synchronized, high-resolution event cameras and standard RGB cameras, we build a hardware synchronized hybrid sensor which combines a high-resolution event camera with a high resolution and high-speed color camera.
We use this hybrid sensor to record a new large-scale dataset which we term the High-Speed Events and RGB (HS-ERGB) dataset 
which we use to validate our video frame interpolation approach. 
The hybrid camera setup is illustrated in Fig.~\ref{fig:dual_setup}.

It features a Prophesee Gen4 (1280$\times$720) event camera (Fig.~\ref{fig:dual_setup} top) and a FLIR BlackFly S global shutter RGB camera (1440$\times$1080) (Fig.~\ref{fig:dual_setup} bottom), 
separated by a baseline of \SI{2.5}{\centi\meter}. Both cameras are hardware synchronized and share a similar field of view (FoV).
We provide a detailed comparison of our setup against the commercially available DAVIS 346~\cite{brandli_2014} and the recently introduced setup~\cite{wang_2020b} in the appendix.
Compared to both~\cite{brandli_2014} and~\cite{wang_2020b} our setup is able to record events at much higher resolution (1280$\times$720 vs. 240$\times$180 or 346$\times$260) and standard frames at much higher framerate (225 FPS vs. 40 FPS or 35 FPS) and with a higher dynamic range (71.45 dB vs. 55 dB or 60 dB).
Moreover, standard frames have a higher resolution compared to the DAVIS sensor (1440$\times$1080 vs. 240$\times$180) and provide color. 
The higher dynamic range and frame rate, enable us to more accurately compare event cameras with standard cameras in highly dynamic scenarios and high dynamic range.
Both cameras are hardware synchronized and aligned via rectification and global alignment. For more synchronization and alignment details see the appendix.

We record data in a variety of conditions, both indoors and outdoors. Sequences were recorded outdoors with exposure times as low as ~\SI{100}{\micro\second} or indoors with exposure times up to ~\SI{1000}{\micro\second}. The dataset features frame rates of 160 FPS, which is much higher than previous datasets, enabling larger frame skips with ground truth color frames. The dataset includes highly dynamic close scenes with non-linear motions and far-away scenes featuring mainly camera ego-motion.
\rev{For far-away scenes, stereo rectification is sufficient for good per-pixel alignment. 
For each sequence, alignment is performed depending on the depth either by stereo rectification or using feature-based homography estimation.To this end, we perform standard stereo calibration between RGB images and E2VID~\cite{rebecq_2019a} reconstructions and rectify the images and events accordingly. For the dynamic close scenes, we additionally estimate a global homography by matching SIFT features~\cite{Lowe04ijcv} between these two images. 
Note that for feature-based alignment to work well, the camera must be static and objects of interest should only move in a fronto-parallel plane at a predetermined depth. While recording we made sure to follow these constraints.}
 
For a more detailed dataset overview we refer to the supplementary material. 


\section{Experiments}
\reviewStepan{All experiments in this work are done using the PyTorch framework~\cite{pytorchurl}. For training, we use the \mbox{Adam} optimizer\cite{Kingma15iclr} with standard settings, batches of size 4 and learning rate $10^4$, which we decrease by a factor of $10$ every 12 epoch. We train each module for 27 epoch. For the training, we use large dataset with synthetic events generated from \textit{Vimeo90k} septuplet dataset~\cite{xue_2019} using the video to events method~\cite{gehrig_2020}, based on the event simulator from \cite{Rebecq18corl}.

We train the network by adding and training
modules one by one, while freezing the weights of all previously trained modules. We train modules in the following order: synthesis-based interpolation, warping-based interpolation, warping refinement, and attention averaging modules. We adopted this training
 because end-to-end training from scratch does not converge, and fine-tuning of the entire network after pretraining only marginally improved the results. We supervise our network with perceptual ~\cite{zhang_2018} and $L^1$ losses as shown in Fig.~\ref{fig:synthesis},~\ref{fig:warp},~\ref{fig:refinement} and~\ref{fig:attention}}. 
 We fine-tune our network on real data module-by-module in the order of training. 
 To measure the quality of interpolated images we use structural similarity (SSIM)~\cite{wang_2004} and peak signal to noise ratio (PSNR) metrics.

\reviewStepan{Note, that the computational complexity of our interpolation method is among the best: on our machine for image resolutions of $640\times 480$, a single interpolation on the GPU takes 878 ms for DAIN~\cite{bao_2019}, 404 ms for BMBC~\cite{park_2020}, 138 ms for ours, 84 ms
for RRIN~\cite{li_2020}, 73 ms for Super SloMo~\cite{jiang_2018} and 33 ms for
LEDVDI~\cite{lin_2020} methods.}

\subsection{Ablation study}

\begin{figure}[htbp]
	\centering
	\includegraphics[width=1\columnwidth]{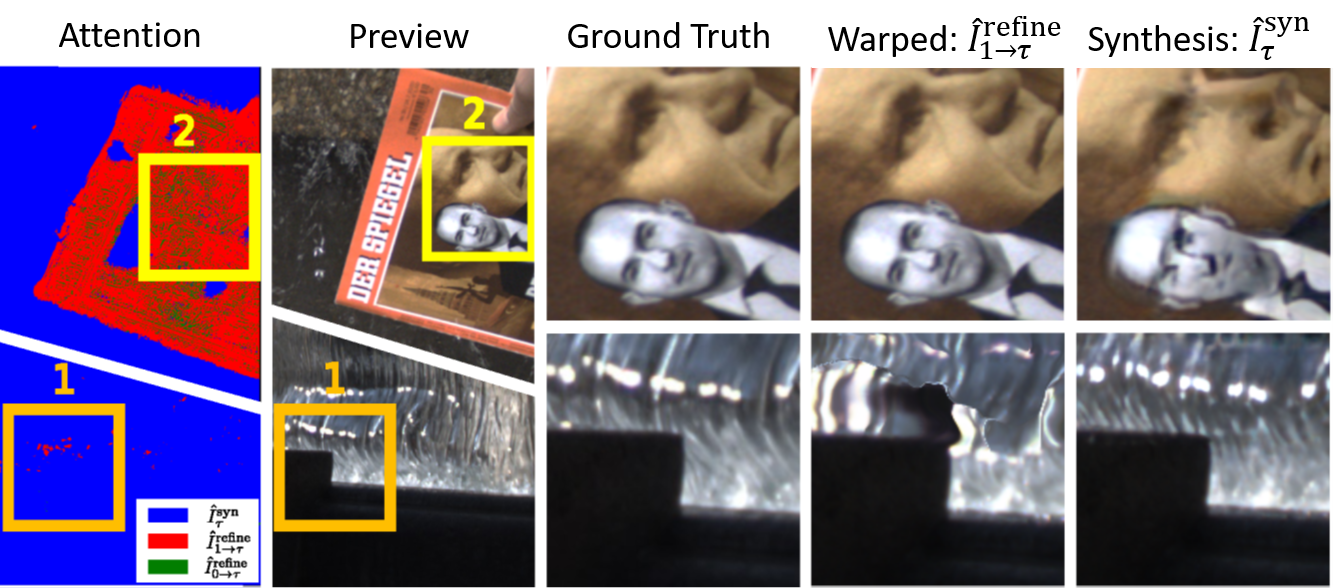}
	\caption{\rev{Complementarity of warping- and synthesis-based interpolation.}}\label{fig:rebuttal}
\end{figure}

To study the contribution of every module of the proposed method to the final interpolation, we investigate the interpolation quality after each module in Fig.~\ref{fig:workflow}, and report their results in Tab.~\ref{tab:ablation}. 
The table shows two notable results. First, it shows that adding a warping refinement block after the simple warping block significantly improves the interpolation result. 
Second, it shows that by attention averaging synthesis-based and warping-based results, the interpolations are improved by 1.7 dB in terms of PSNR. 
\rev{This is because the attention averaging module combines the advantages of both methods. To highlight this further, we illustrate example reconstructions from these two modules in Fig.~\ref{fig:rebuttal}.
As can be seen, the warping-based module excels at reconstructing textures in non-occluded areas (fourth column) while the synthesis module performs better in regions with difficult lighting conditions (fifth column). 
The attention module successfully combines the best parts of both modules (first column).}

\begin{table}[htbp]
  \caption{Quality of interpolation after each module on Vimeo90k~(denoising) validation set. For SSIM and PSNR we show mean and one standard deviation. The best result is highlighted.
  }\label{tab:ablation}
\vspace{0.5em}
\centering
\small
\begin{tabularx}{\columnwidth}{ llR }
\hline
\textbf{Module} & \textbf{PSNR} & \textbf{SSIM} \\
\hline
Warping interpolation  & 26.68$\pm$3.68 & 0.926$\pm$0.041 \\
Interpolation by synthesis & 34.10$\pm$3.98 & 0.964$\pm$0.029 \\
Warping refinement & 33.02$\pm$3.76 &  0.963$\pm$0.026 \\
\textbf{Attention averaging (ours)} & \textbf{35.83$\pm$3.70} & \textbf{0.976$\pm$0.019} \\
\hline
\end{tabularx}
\end{table}

\subsection{Benchmarking}\label{sec:benchmarking}

\textbf{Synthetic datasets.} We compare the proposed method, which we call \textit{Time Lens}, to four state-of-the-art frame-based interpolation methods \textit{DAIN}~\cite{bao_2019}, \textit{RRIN}~\cite{li_2020}, \textit{BMBC}~\cite{park_2020}, \textit{SuperSloMo}~\cite{jiang_2018}, event-based video reconstruction method \textit{E2VID}~\cite{rebecq_2019b} and two event and frame-based methods~
EDI~\cite{pan_2019} and LEDVDI~\cite{lin_2020} on popular video interpolation benchmark datasets, such as \textit{Vimeo90k~(interpolation)}~\cite{xue_2019}, \textit{Middlebury}~\cite{baker_2011}. During the evaluation, we take original video sequence, skip 1 or 3 frames respectively, reconstruct them using interpolation method and compare to ground truth skipped frames. Events for event-based methods we simulate using~\cite{gehrig_2020} from the skipped frames. We do not fine-tune the methods for each dataset but simply use pre-trained models provided by the authors. We summarise the results in Tab.~\ref{tab:benchmarking_synthetic}.

As we can see, the proposed method outperforms other method across datasets in terms of average PSNR (up to 8.82 dB improvement) and SSIM scores (up to 0.192 improvement). 
As before these improvements stem from the use of auxiliary events during the prediction stage which allow our method to perform accurate frame interpolation, event for very large non-linear motions.
Also, it has significantly lower standard deviation of the PSNR (2.53 dB vs. 4.96 dB) and SSIM (0.025 vs. 0.112) scores, which suggests more consistent performance across examples. Also, we can see that PSNR and SSIM scores of the proposed method degrades to much lesser degree than scores of the frame-based methods (up to 1.6 dB vs. up to 5.4 dB), as we skip and attempt to reconstruct more frames. This suggests that our method is more robust to non-linear motion than frame-based methods.
  
\begin{table*}[tb]
  \caption{Results on standard video interpolation benchmarks such as \textit{Middlebury}~\cite{baker_2011}, \textit{Vimeo90k}~(interpolation)~\cite{xue_2019} and \textit{GoPro}~\cite{nah_2017}. In all cases, we use a test subset of the datasets. To compute SSIM and PSNR, we downsample the original video and reconstruct the skipped frames. 
  For Middlebury and Vimeo90k~(interpolation), we skip 1 and 3 frames, and for GoPro we skip 7 and 15 frames due its its high frame rate of 240 FPS. \textit{Uses frames} and \textit{Uses events} indicate if a method uses frames and events for interpolation. For event-based methods we generate events from the skipped frames using the event simulator~\cite{gehrig_2020}. \textit{Color} indicates if a method works with color frames. For SSIM and PSNR we show mean and one standard deviation. Note, that we can not produce results with 3 skips on the Vimeo90k dataset, since it consists of frame triplet. We show the best result in each column in bold and the second-best using underscore text.
  }\label{tab:benchmarking_synthetic}
\vspace{0.5em}
\resizebox{.95\textwidth}{!}{
\centering
\small
\begin{tabularx}{\textwidth}{ X ccc XXXX}
\hline
\textbf{Method} & \textbf{Uses frames} & \textbf{Uses events} & \textbf{Color} & \textbf{PSNR} & \textbf{SSIM} & \textbf{PSNR} & \textbf{SSIM} \\
\hline
\multicolumn{4}{l}{\textbf{Middlebury}~\cite{baker_2011}}  & \multicolumn{2}{c}{\textbf{1 frame skip}} & \multicolumn{2}{c}{\textbf{3 frames skips}} \\
\hline
\Tstrut DAIN~\cite{bao_2019} & \greencheck & \redcross & \greencheck & 30.87$\pm$5.38  & \underline{0.899$\pm$0.110} & 26.67$\pm$4.53 & \underline{0.838$\pm$0.130} \\
SuperSloMo~\cite{jiang_2018}& \greencheck & \redcross & \greencheck & 29.75$\pm$5.35 & 0.880$\pm$0.112  & 26.43$\pm$5.30  &  0.823$\pm$0.141 \\
RRIN~\cite{li_2020} & \greencheck & \redcross & \greencheck & \underline{31.08$\pm$5.55} & 0.896$\pm$0.112
& \underline{27.18$\pm$5.57}
& 0.837$\pm$0.142 \\
BMBC~\cite{park_2020} & \greencheck & \redcross & \greencheck & 30.83$\pm$6.01 & 0.897$\pm$0.111 & 26.86$\pm$5.82 &  0.834$\pm$0.144   \\
E2VID~\cite{rebecq_2019a} & \redcross & \greencheck & \redcross & 11.26$\pm$2.82 & 0.427$\pm$0.184 & 26.86$\pm$5.82 &  0.834$\pm$0.144   \\
EDI~\cite{pan_2019}&\greencheck&\greencheck&\redcross&19.72$\pm$2.95&0.725$\pm$0.155&18.44$\pm$2.52&0.669$\pm$0.173\\
\mbox{\textbf{Time Lens~(ours)}} & \greencheck & \greencheck & \greencheck & \textbf{33.27$\pm$3.11} & \textbf{0.929$\pm$0.027} & \textbf{32.13$\pm$2.81} & \textbf{0.908$\pm$0.039}\\
\hline
\multicolumn{4}{l}{\textbf{Vimeo90k~(interpolation)}~\cite{xue_2019}}  & \multicolumn{2}{c}{\textbf{1 frame skip}} & \multicolumn{2}{c}{\textbf{3 frames skips}} \\
\hline
\Tstrut DAIN~\cite{bao_2019} & \greencheck & \redcross & \greencheck & 34.20$\pm$4.43  & 0.962$\pm$0.023 & - & - \\
SuperSloMo~\cite{jiang_2018}& \greencheck & \redcross & \greencheck & 32.93$\pm$4.23 & 0.948$\pm$0.035  & -  &  - \\
RRIN~\cite{li_2020} & \greencheck & \redcross & \greencheck & \underline{34.72$\pm$4.40} & 0.962$\pm$0.029 & - & - \\
BMBC~\cite{park_2020} & \greencheck & \redcross & \greencheck & 34.56$\pm$4.40 & \underline{0.962$\pm$0.024} & - &  -   \\
E2VID~\cite{rebecq_2019a} & \redcross & \greencheck & \redcross & 10.08$\pm$2.89 & 0.395$\pm$0.141 & - &  -   \\
EDI~\cite{pan_2019} & \greencheck & \greencheck & \redcross & 20.74$\pm$3.31 & 0.748$\pm$0.140 & - &  -   \\
\mbox{\textbf{Time Lens~(ours)}} & \greencheck & \greencheck & \greencheck & \textbf{36.31$\pm$3.11} & \textbf{0.962$\pm$0.024} & - & -\\
\hline
\multicolumn{4}{l}{\textbf{GoPro}~\cite{nah_2017}}  & \multicolumn{2}{c}{\textbf{7 frames skip}} & \multicolumn{2}{c}{\textbf{15 frames skips}} \\
\hline
\Tstrut DAIN~\cite{bao_2019} & \greencheck & \redcross & \greencheck & 28.81$\pm$4.20  & 0.876$\pm$0.117 &  \underline{24.39$\pm$4.69} & 0.736$\pm$0.173 \\
SuperSloMo~\cite{jiang_2018}& \greencheck & \redcross & \greencheck & 28.98$\pm$4.30 & 0.875$\pm$0.118  & 24.38$\pm$4.78  &  0.747$\pm$0.177 \\
RRIN~\cite{li_2020} & \greencheck & \redcross & \greencheck & 28.96$\pm$4.38 & \underline{0.876$\pm$0.119} & 24.32$\pm$4.80 & \underline{0.749$\pm$0.175} \\
BMBC~\cite{park_2020} & \greencheck & \redcross & \greencheck & \underline{29.08$\pm$4.58} & 0.875$\pm$0.120 & 23.68$\pm$4.69 &  0.736$\pm$0.174   \\
E2VID~\cite{rebecq_2019a} & \redcross & \greencheck & \redcross & 9.74$\pm$2.11 & 0.549$\pm$0.094 & 9.75$\pm$2.11 &  0.549$\pm$0.094 \\
EDI~\cite{pan_2019}&\greencheck&\greencheck&\redcross&18.79$\pm$2.03&0.670$\pm$0.144&17.45$\pm$2.23&0.603$\pm$0.149\\
\mbox{\textbf{Time~Lens~(ours)}} & \greencheck & \greencheck & \greencheck & \textbf{34.81$\pm$1.63} & \textbf{0.959$\pm$0.012} & \textbf{33.21$\pm$2.00} & \textbf{0.942$\pm$0.023} \\
\hline
\end{tabularx}
}
\end{table*}

\textbf{High Quality Frames~(HQF) dataset.} We also evaluate our method on \textit{High Quality Frames (HQF)} dataset~\cite{stoffregen_2020} collected using DAVIS240 event camera that consists of video sequences without blur and saturation. During evaluation, we use the same methodology as for the synthetic datasets, with the only difference that in this case we use real events. In the evaluation, we consider two versions of our method: \textit{Time Lens-syn}, which we trained only on synthetic data, and \textit{Time Lens-real}, which we trained on synthetic data and fine-tuned on real event data from our own DAVIS346 camera. We summarise our results in Tab.~\ref{tab:benchmarking_davis}.

The results on the dataset are consistent with the results on the synthetic
datasets: the proposed method outperforms state-of-the-art frame-based methods
and produces more consistent results over examples. As we increase the number of
frames that we skip, the performance gap between the proposed method and the
other methods widens from 2.53 dB to 4.25 dB, also the results of other methods become less consistent
which is reflected in higher deviation of PSNR and SSIM scores. 
For a more detailed discussion about the impact of frame skip length and performance, see the appendix.
Interestingly, fine-tuning of the proposed method on real event data, captured by another camera, greatly boosts the performance of our method by an average of 1.94 dB. This suggest that existence of large domain gap between synthetic and real event data.

\begin{figure*}[htbp]
	\centering
\animategraphics[width=0.35\columnwidth,autoplay,loop]{5}{figures/fountain/fountain_}{1}{20}%
	\includegraphics[width=0.67\columnwidth]{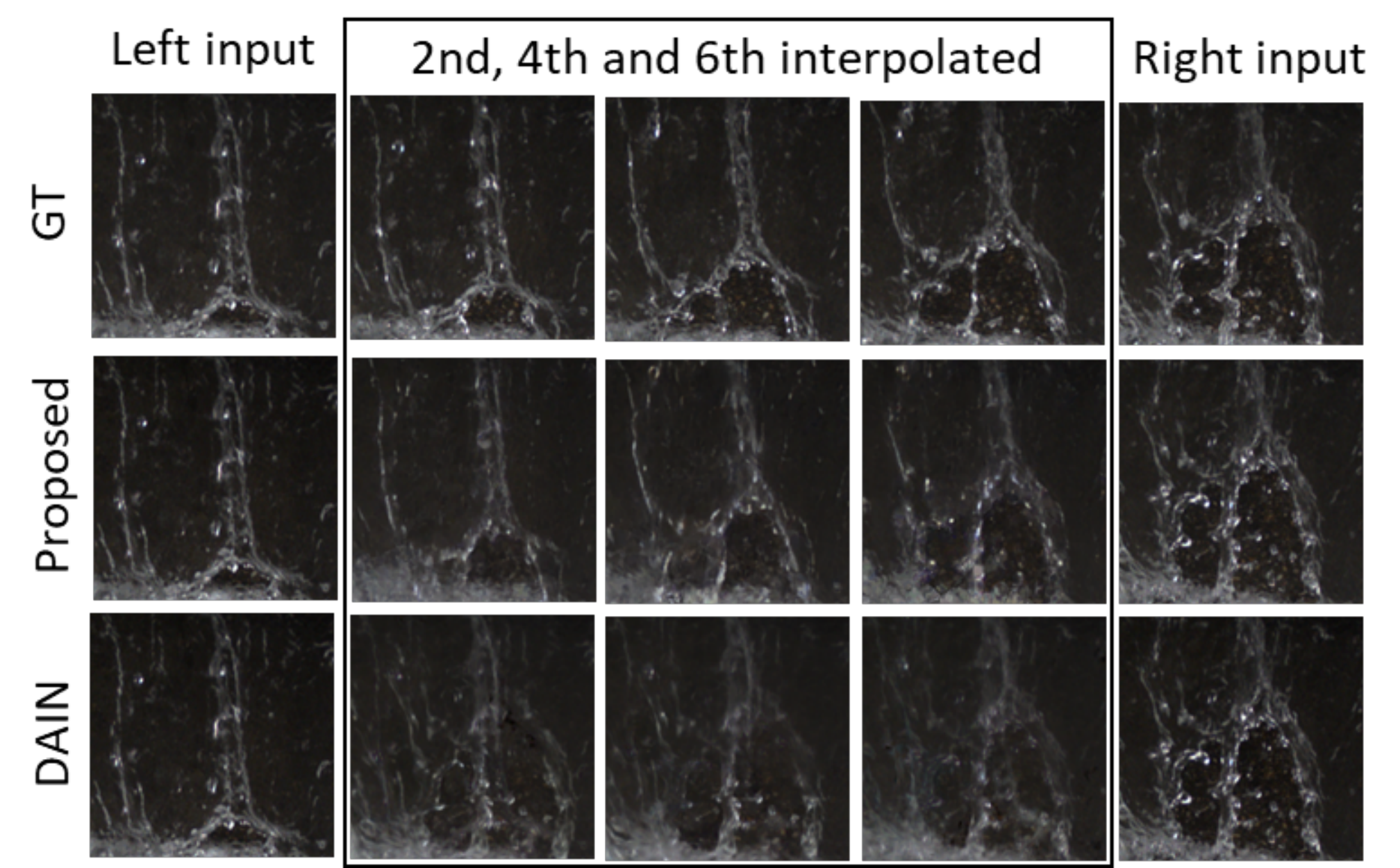}%
	\animategraphics[width=0.41\columnwidth,autoplay,loop]{5}{figures/waterbomb/waterbomb_}{1}{1}
	\includegraphics[width=0.66\columnwidth]{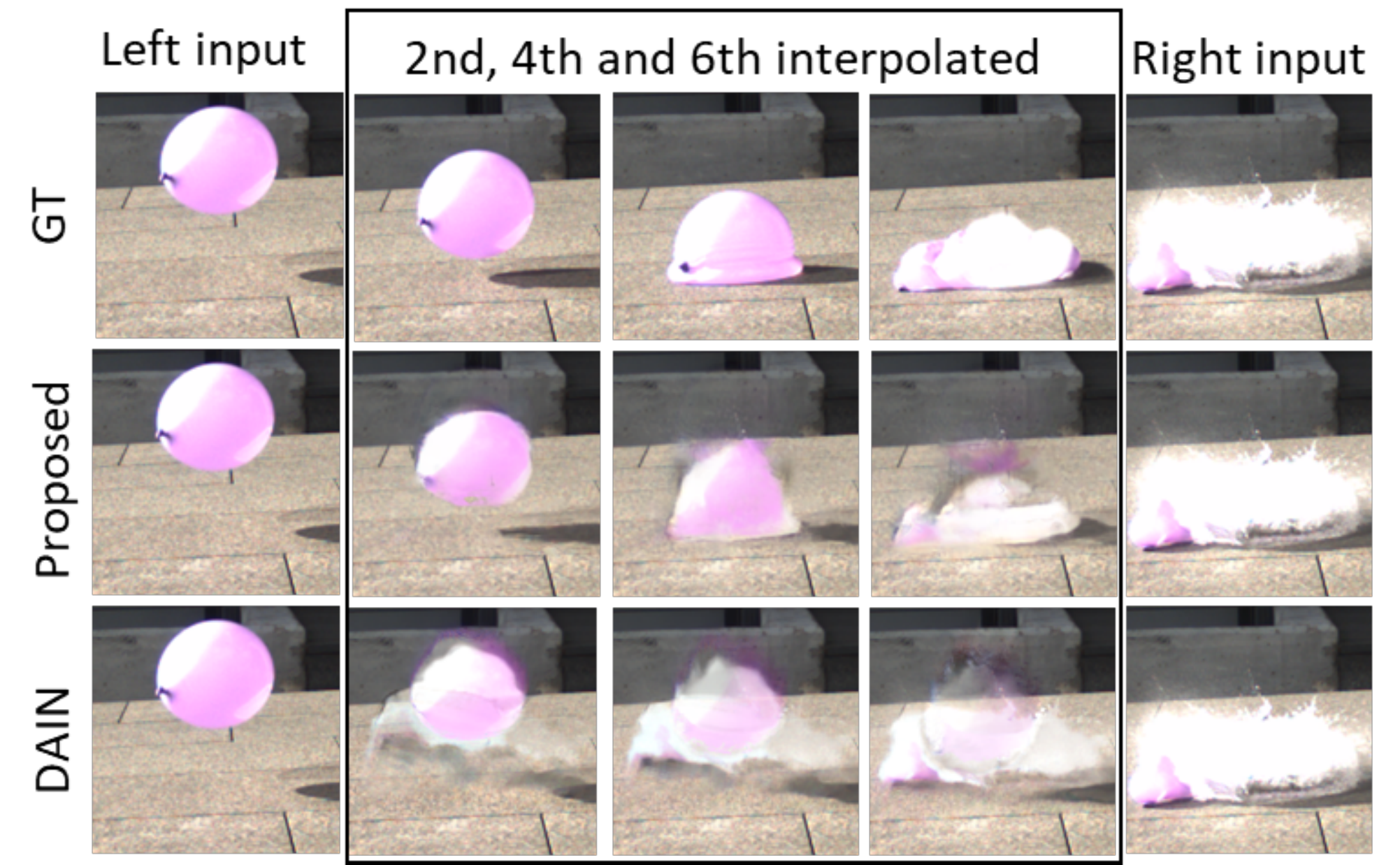}\\
	\animategraphics[width=0.46\columnwidth,autoplay,loop]{5}{figures/fountain_2/fountain_2_}{0}{0}
	\includegraphics[width=0.635\columnwidth]{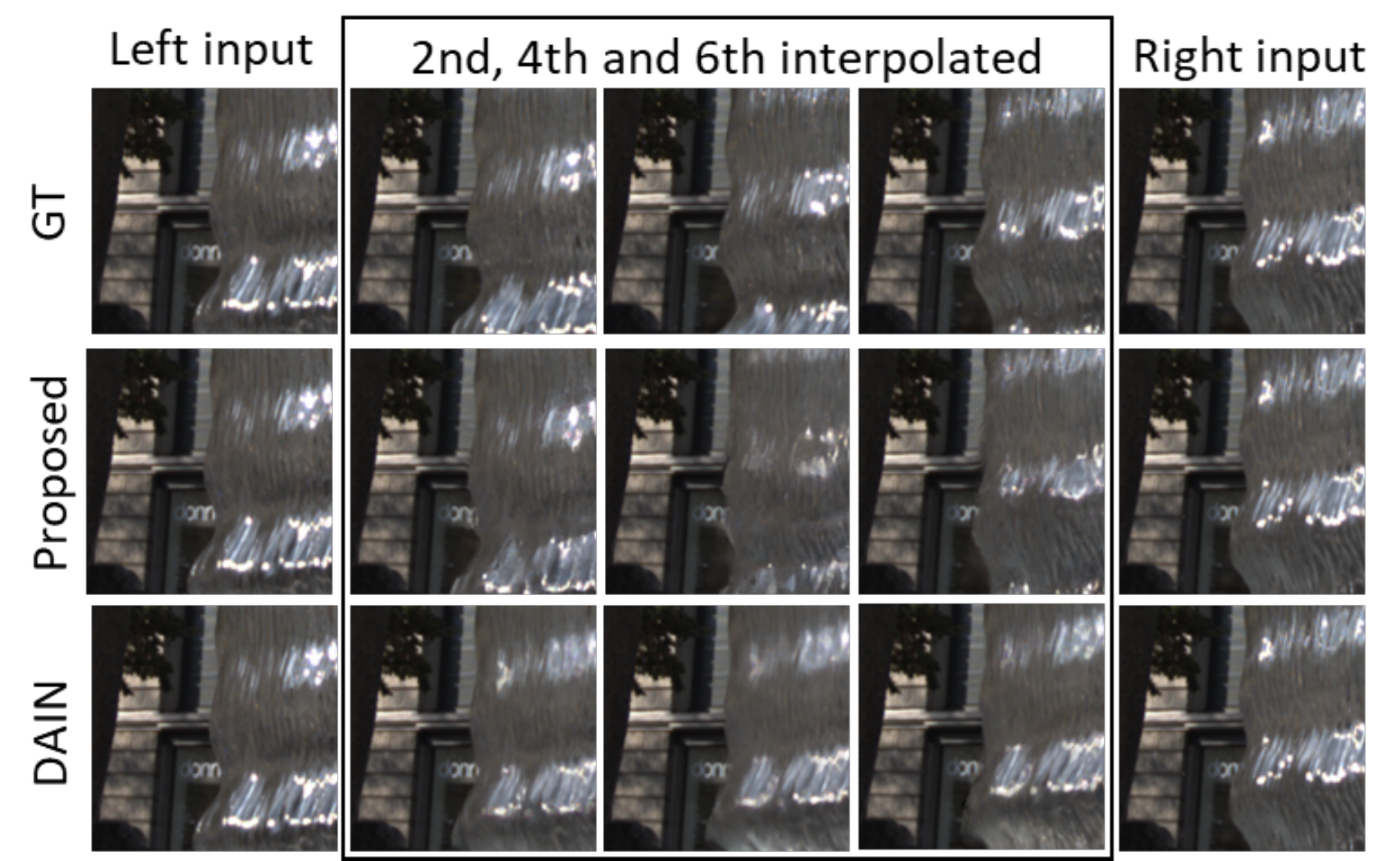}%
    \animategraphics[width=0.385\columnwidth,autoplay,loop]{5}{figures/umbrella/umbrella_}{1}{1}
	\includegraphics[width=0.635\columnwidth]{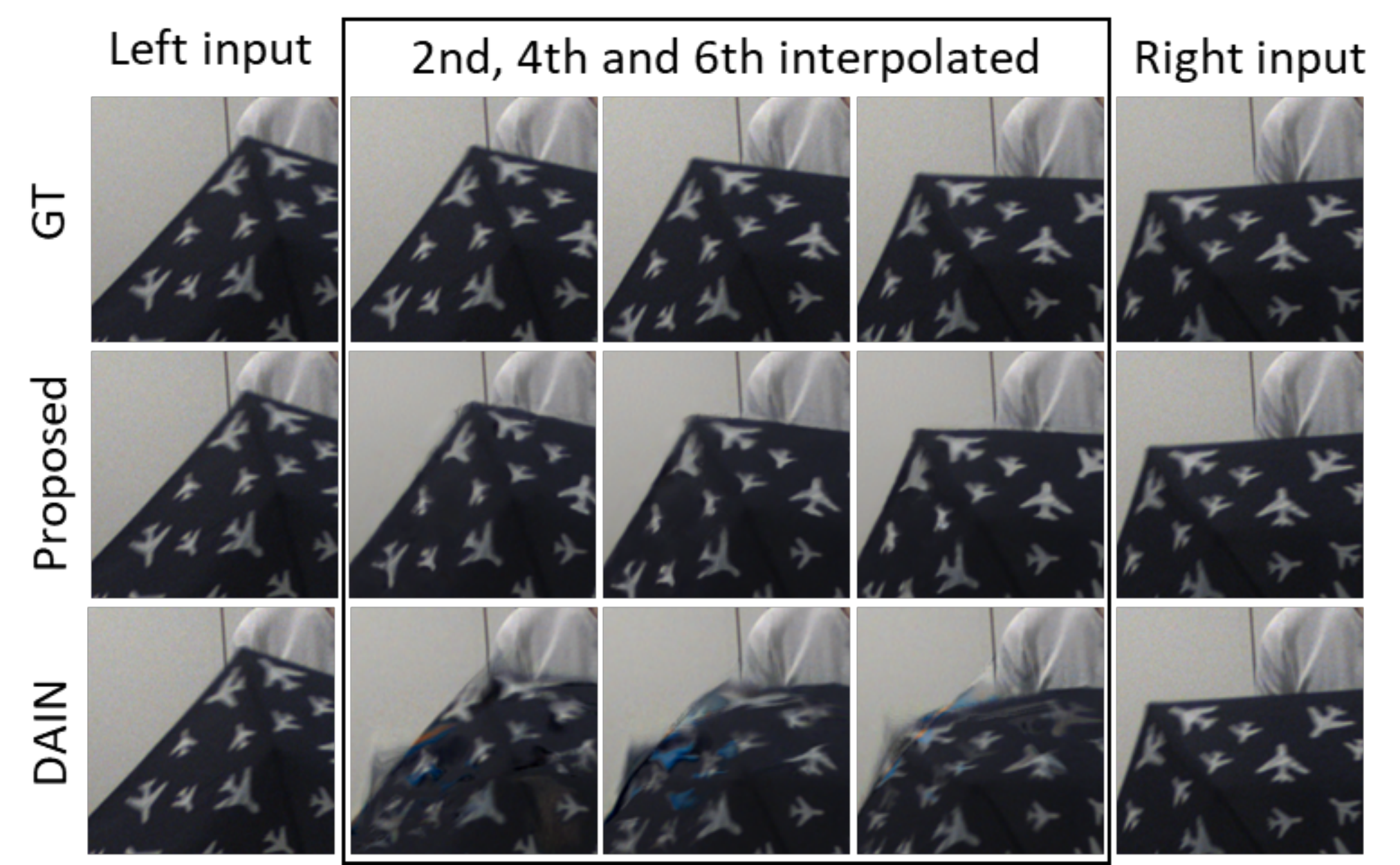}\\

\caption{Qualitative results for the proposed method and its closes competitor DAIN~\cite{bao_2019} on our Dual Event and Color Camera Dataset test sequences: ``Fountain Schaffhauserplatz''~(top-left), ``Fountain Bellevue''~(bottom-left) ``Water bomb''~(top-right) and ``Umbrella"~(bottom-right). For each sequence, the figure shows interpolation results on the left~(the animation can be viewed in Acrobat Reader) and close-up interpolation results on the right. The close-ups, show input left and right frame and intermediate interpolated frames. }
	\label{fig:qualitative}
\end{figure*}

\textbf{High Speed Event-RGB dataset.} Finally, we evaluate our method on our dataset introduced in~\S~\ref{sec:high_speed_events_and_rgb}. As clear from Tab.~\ref{tab:benchmarking_decc}, our method, again significantly outperforms frame-based and frame-plus-event-based competitors. In Fig.~\ref{fig:qualitative} we show several examples from the HS-ERGB test set which show that, compared to competing frame-based method, our method can interpolate frames in the case of nonlinear~(``Umbrella'' sequence) and non-rigid motion~(``Water Bomb''), and also handle illumination changes~(``Fountain Schaffhauserplatz'' and ``Fountain Bellevue''). 

\begin{table*}[tb]
  \caption{Benchmarking on the High Quality Frames~(HQF) DAVIS240 dataset. We do not fine-tune our method and other methods and use models provided by the authors. We evaluate methods on all sequences of the dataset. To compute SSIM and PSNR, we downsample the original video by skip 1 and 3 frames, reconstruct these frames and compare them to the skipped frames. In \textit{Uses frames} and \textit{Uses events} columns we specify if a method uses frames and events for interpolation. In the \textit{Color} column, we indicate if a method works with color frames. In the table, we present two versions of our method: \textit{Time~Lens-syn}, which we trained only on synthetic data, and \textit{Time~Lens-real}, which we trained on synthetic data and fine-tuned on real event data from our own DAVIS346 camera. For SSIM and PSNR, we show mean and one standard deviation. We show the best result in each column in bold and the second-best using underscore text.
  }\label{tab:benchmarking_davis}
\vspace{0.5em}
\centering
\small
\begin{tabularx}{\textwidth}{ X ccc XXXX}
\hline
\textbf{Method} & \textbf{Uses frames} & \textbf{Uses events} & \textbf{Color} & \textbf{PSNR} & \textbf{SSIM} & \textbf{PSNR} & \textbf{SSIM} \\
\hline
\multicolumn{4}{l}{} & \multicolumn{2}{c}{\textbf{1 frame skip}} & \multicolumn{2}{c}{\textbf{3 frames skips}} \\
\hline
\Tstrut DAIN~\cite{bao_2019} & \greencheck & \redcross & \greencheck & 29.82$\pm$6.91  & 0.875$\pm$0.124 & 26.10$\pm$7.52 & \underline{0.782$\pm$0.185} \\
SuperSloMo~\cite{jiang_2018}& \greencheck & \redcross & \greencheck & 28.76$\pm$6.13 & 0.861$\pm$0.132  & 25.54$\pm$7.13  & 0.761$\pm$0.204 \\
RRIN~\cite{li_2020} & \greencheck & \redcross & \greencheck & 29.76$\pm$7.15 & 0.874$\pm$0.132 & 26.11$\pm$7.84 & 0.778$\pm$0.200\\
BMBC~\cite{park_2020} & \greencheck & \redcross & \greencheck & \underline{29.96$\pm$7.00} & \underline{0.875$\pm$0.126} & \underline{26.32$\pm$7.78} & 0.781$\pm$0.193 \\
E2VID~\cite{rebecq_2019a} & \redcross & \greencheck & \redcross & 6.70$\pm$2.19 & 0.315$\pm$0.124 & 6.70$\pm$2.20 & 0.315$\pm$0.124     \\
EDI~\cite{pan_2019} & \greencheck & \greencheck & \redcross & 18.7$\pm$6.53 & 0.574$\pm$0.244 & 18.8$\pm$6.88 & 0.579$\pm$0.274     \\
\hline
\mbox{\textbf{Time~Lens-syn~(our)}} & \greencheck & \greencheck & \greencheck & 30.57$\pm$5.01 & 0.903$\pm$0.067 & 28.98$\pm$5.09  & 0.873$\pm$0.086 \\
\mbox{\textbf{Time~Lens-real~(ours)}} & \greencheck & \greencheck & \greencheck & \textbf{32.49$\pm$4.60} & \textbf{0.927$\pm$0.048} & \textbf{30.57$\pm$5.08}  & \textbf{0.900$\pm$0.069}\\
\hline
\vspace{1ex}
\end{tabularx}
\end{table*}

\begin{table*}[tb]
  \caption{Benchmarking on the test set of the High Speed Event and RGB camera (HS-ERGB) dataset. We report PSNR and SSIM for all sequences by skipping 5 and 7 frames respectively, and reconstructing the missing frames with each method. By design LEDVDI~\cite{lin_2020} can interpolate only 5 frames. \textit{Uses frames} and \textit{Uses events} indicate if a method uses frames or events respectively. \textit{Color} indicates whether a method works with color frames. For SSIM and PSNR the scores are averaged over the sequences. Best results are shown in bold and the second best are underlined.}\label{tab:benchmarking_decc}
\vspace{0.5em}
\centering
\small
\begin{tabularx}{\textwidth}{ X ccc XXXX}
\hline
\textbf{Method} & \textbf{Uses frames} & \textbf{Uses events} & \textbf{Color} & \textbf{PSNR} & \textbf{SSIM} & \textbf{PSNR} & \textbf{SSIM} \\
\hline
\multicolumn{4}{l}{\textbf{Far-away sequences}}  & \multicolumn{2}{c}{\textbf{5 frame skip}} & \multicolumn{2}{c}{\textbf{7 frames skips}} \\
\hline
\Tstrut DAIN~\cite{bao_2019} & \greencheck & \redcross & \greencheck & \underline{27.92$\pm$1.55} & \underline{0.780$\pm$0.141} & \underline{27.13$\pm$1.75} & \underline{0.748$\pm$0.151} \\
SuperSloMo~\cite{jiang_2018}& \greencheck & \redcross & \greencheck & 25.66$\pm$6.24 & 0.727$\pm$0.221 & 24.16$\pm$5.20 &  0.692$\pm$0.199 \\
RRIN~\cite{li_2020}         & \greencheck & \redcross & \greencheck & 25.26$\pm$5.81 & 0.738$\pm$0.196 & 23.73$\pm$4.74 & 0.703$\pm$0.170\\
BMBC~\cite{park_2020}       & \greencheck & \redcross & \greencheck &   25.62$\pm$6.13 & 0.742$\pm$0.202 & 24.13$\pm$4.99 & 0.710$\pm$0.175\\
LEDVDI~\cite{lin_2020} & \greencheck & \greencheck & \redcross & 12.50$\pm$1.74 & 0.393$\pm$0.174 & n/a & n/a \\
\mbox{\textbf{Time~Lens~(ours)}} & \greencheck & \greencheck & \greencheck & \textbf{33.13$\pm$2.10} & \textbf{0.877$\pm$0.092} & \textbf{32.31$\pm$2.27} & \textbf{0.869$\pm$0.110}\\
\hline
\multicolumn{4}{l}{\textbf{Close planar sequences}}  & \multicolumn{2}{c}{\textbf{5 frame skip}} & \multicolumn{2}{c}{\textbf{7 frames skips}} \\
\hline
\Tstrut DAIN~\cite{bao_2019} & \greencheck & \redcross & \greencheck & 29.03$\pm$4.47 & 0.807$\pm$0.093 & \underline{28.50$\pm$4.54} & $0.801\pm0.096$ \\
SuperSloMo~\cite{jiang_2018}& \greencheck & \redcross & \greencheck & 28.35$\pm$4.26 & 0.788$\pm$0.098 & 27.27$\pm$4.26 &  $0.775\pm0.099$ \\
RRIN~\cite{li_2020}         & \greencheck & \redcross & \greencheck & 28.69$\pm$4.17 & 0.813$\pm$0.083 & 27.46$\pm$4.24 & 0.800$\pm$0.084\\
BMBC~\cite{park_2020}       & \greencheck & \redcross & \greencheck &   \underline{29.22$\pm$4.45} & \underline{0.820$\pm$0.085} & 27.99$\pm$4.55 & \underline{0.808$\pm$0.084} \\
LEDVDI~\cite{lin_2020}       & \greencheck & \greencheck & \redcross & 19.46$\pm$4.09 & 0.602$\pm$0.164 & n/a & n/a \\
\mbox{\textbf{Time~Lens~(ours)}} & \greencheck & \greencheck & \greencheck & \textbf{32.19$\pm$4.19}  & \textbf{0.839$\pm$0.090} & \textbf{31.68$\pm$4.18}  & \textbf{0.835$\pm$0.091}\\
\hline
\end{tabularx}
\end{table*}

\section{Conclusion}
In this work, we introduce Time Lens, a method that can show us what happens in the blind-time between two intensity frames using high temporal resolution information from an event camera. It works by leveraging the advantages of synthesis-based approaches, which can handle changing illumination conditions and non-rigid motions, and flow-based approach, relying on motion estimation from events. It is therefore robust to motion blur and non-linear motions. The proposed method achieves an up to 5.21 dB improvement over state-of-the-art frame-based and event-plus-frames-based methods on both synthetic and real datasets. 
In addition, we release the first High Speed Event and RGB (HS-ERGB) dataset, which aims at pushing the limits of existing interpolation approaches by establishing a new benchmark for both event- and frame-based video frame interpolation methods. 

\\
\section{Acknowledgement}
\rev{This work was supported by Huawei Zurich Research Center; by the National Centre of Competence in Research~(NCCR) Robotics through the Swiss National Science Foundation~(SNSF); the European Research Council~(ERC) under the European Union’s  Horizon 2020 research and innovation programme~(Grant agreement No. 864042). }

\section{Video Demonstration} 
This PDF is accompanied with a video showing advantages of the proposed method compared to state-of-the-art frame-based methods published over recent months, as well as potential practical applications of the method.  
 
\section{Backbone network architecture}

\begin{figure}[htbp]
	\centering
	\includegraphics[width=0.99\columnwidth]{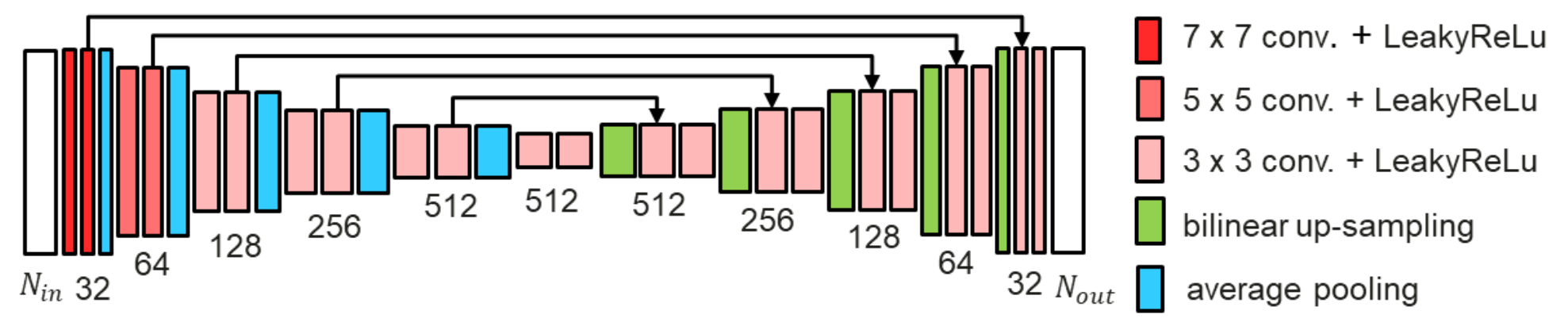}
	\caption{Backbone hourglass network that we use in all modules of the proposed method.}
	\label{fig:backbone}
\end{figure}%

For all modules in the proposed method, we use the same backbone architecture which is an \textit{hourglass network} with shortcut connections between the contracting and the expanding parts similar to~\cite{jiang_2018} which we show in Fig.~\ref{fig:backbone}.

\section{Additional Ablation Experiments}

\begin{figure}[htbp]
	\centering
	\includegraphics[width=0.99\columnwidth]{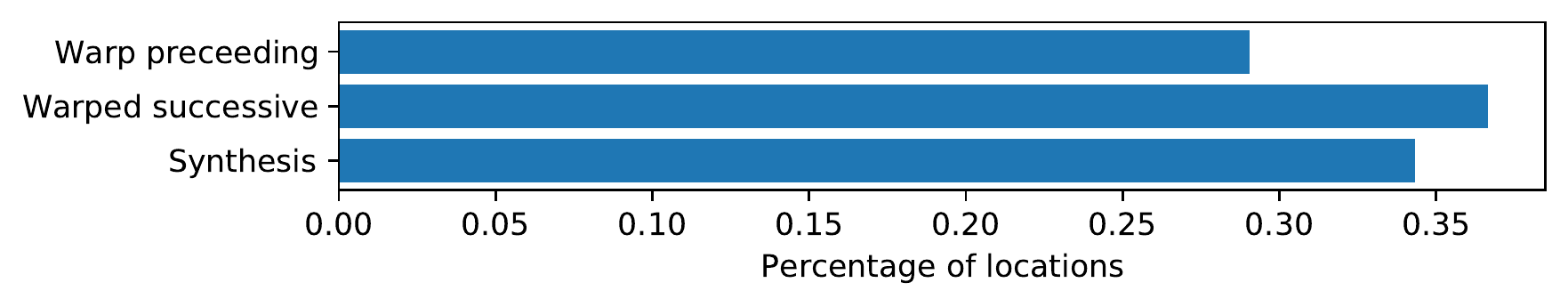}
	\caption{Percentage of pixels each interpolation method contributes on average to the final interpolation result for Vimeo90k~(denoising) validation set. Note, that all methods contribute almost equally to the final result and thus are equally important.}
	\label{fig:distribution}
\end{figure}%

\begin{table}[htbp]
  \caption{Importance of inter-frame events on Middlebury test set. To compute SSIM and PSNR, we skip one frame of the original video, reconstruct it and compare to the skipped frame. One version of the proposed method has access to the events synthesized from the skipped frame and another version does not have inter-frame information. We also show performance of frame-based SuperSloMo method~\cite{jiang_2018}, that is used in event simulator for reference. We highlight the best performing method.}\label{tab:inter_frame_events}
\vspace{0.5em}
\centering
\small
\begin{tabularx}{\columnwidth}{ llR }
\hline
\textbf{Method} & \textbf{PSNR} & \textbf{SSIM} \\
\hline
\textbf{With inter-frame events (ours)} & \textbf{33.27$\pm$3.11} & \textbf{0.929$\pm$0.027} \\
Without inter-frame events & 29.03$\pm$4.85 & 0.866$\pm$0.111 \\
\hline
SuperSloMo~\cite{jiang_2018}& 29.75$\pm$5.35 & 0.880$\pm$0.112  \\
\hline
\end{tabularx}
\end{table}

\textbf{Importance of inter-frame events}. To study the importance of additional information provided by events, we skip every second frame of the original video and attempt to reconstruct it using two versions of the proposed method. 
One version has access to the events synthesized from the skipped frame and another version does not have inter-frame information. As we can see from the Tab.~\ref{tab:inter_frame_events}, the former significantly outperforms the later by a margin of 4.24dB. 
Indeed this large improvements can be explained by the fact that the method with inter-frame events has implicit access to the ground truth image it tries to reconstruct, albeit in the form of asynchronous events. 
This highlights that our network is able to efficiently decode the asynchronous intermediate events to recover  the missing frame. 
Moreover, this shows that the addition of events has a significant impact on the final task performance, proving the usefulness of an event camera as an auxiliary sensor.

\textbf{Importance of each interpolation method.} To study relative importance of \textit{synthesis-based} and \textit{warping-based} interpolation methods, we compute the percentage of pixels that each method contribute on average to the final interpolation result for the Vimeo90k~(denoising) validation dataset and show the result in Fig.~\ref{fig:distribution}. As it is clear from the figure, all the methods contribute almost equally to the final result and thus are all equally important.  

\begin{figure}[htbp]
	\centering
	\includegraphics[width=0.9\columnwidth]{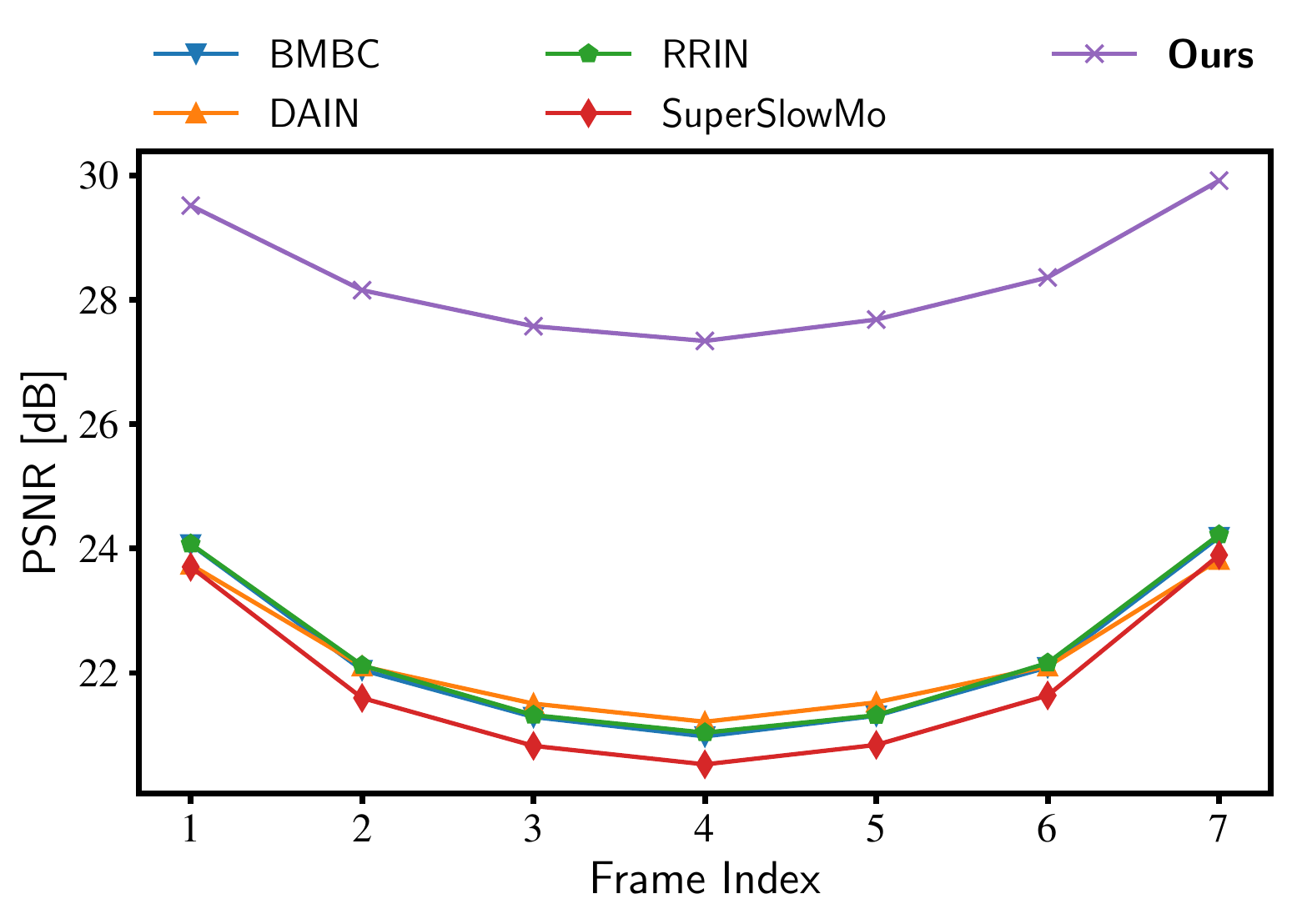}
    \caption{``Rope plot'' showing interpolation quality as a function of distance from input boundary frames on High Quality Frames dataset. We skip all but every 7th frame and restore them using events and remaining frames. For each skip position, we compute average PSNR of the restored frame over entire dataset. We do not fine-tune the proposed and competing methods on the HQF dataset and simply use pre-trained models provided by the authors. Note, that the proposed method have the highest PSNR. Also, its PSNR decreases much slower than PSNR of other methods we move away from the input boundary frames.}
	\label{fig:rope_plot}
\end{figure}

\textbf{``Rope'' plot.} To study how the interpolation quality decreases with the distance to the input frames, we skip all but every 7th frame in the input videos from the High Quality Frames dataset, restore them using our method and compare to the original frames. For each skipped frame position, we compute average PSNR of the restored frame over entire dataset and show results in Fig.~\ref{fig:rope_plot}. As clear from the figure, the proposed method has the highest PSNR. Also, its PSNR decreases much slower than PSNR of the competing methods as we move away from the boundary frames. 

\section{Additional Benchmarking Results}

\begin{table*}[htbp]
  \caption{Results on High Quality Frames~\cite{stoffregen_2020} with fine-tuning. Due to the time limitations, we only fine-tuned the proposed method and RRIN~\cite{li_2020} method, that performed well across synthetic and real datasets. For evaluation, we used ``poster\_pillar\_1'', ``slow\_and\_fast\_desk'', ``bike\_bay\_hdr'' and ``desk'' sequences of the set and other sequences we used for the fine-tuning. For SSIM and PSNR, we show mean and one standard deviation across frames of all sequences.}\label{tab:finetuning}
\vspace{0.5em}
\centering
\begin{tabularx}{\textwidth}{ l XXXX}
\hline
\multirow{2}{*}{\textbf{Method} } & \multicolumn{2}{c}{\textbf{1 skip}} & \multicolumn{2}{c}{\textbf{3 skips}} \\
 & \textbf{PSNR} & \textbf{SSIM} & \textbf{PSNR} & \textbf{SSIM} \\
\hline
RRIN~\cite{li_2020} & 28.62$\pm$5.51 & 0.839$\pm$0.132 & 25.36$\pm$5.70 & 0.750$\pm$0.173 \\
\mbox{\textbf{Time Lens~(Ours)}} & \textbf{33.42$\pm$3.18} & \textbf{0.934$\pm$0.041} & \textbf{32.27$\pm$3.44} & \textbf{0.917$\pm$0.054}\\
\hline
\end{tabularx}
\end{table*}%
\begin{table*}[tb]
\caption{Comparison of our HS-ERGB dataset against publicly available High Quality Frames~(HQF) dataset, acquired by DAVIS 346~\cite{brandli_2014} and Guided~Event~Filtering~(GEF) dataset, acquired by setup with DAVIS240 and RGB camera mounted with beam splitter~ \cite{wang_2020b}. Note, that in contrast to the previous datasets, the proposed dataset has high resolution of event data, and high frame rate. Also, it is the first dataset acquired by dual system with event and frame sensors arranged in stereo configuration.}\label{tab:dual_setup_specs}
\centering
\resizebox{1\textwidth}{!}{%
\begin{tabular}{l | llll | ll | cc}
\hline
& \multicolumn{4}{c}{Frames} & \multicolumn{2}{c}{Events} &  &  \\ 
& FPS & Dynamic Range,~[dB] & Resolution & Color & Dynamic~Range,~dB & Resolution  & Sync. & Aligned \\
\hline
DAVIS 346 \cite{brandli_2014} & 40 & 55  & 346 $\times$ 260 & \redcross & 120 & 346 $\times$ 260 & \greencheck                             & \greencheck                        \\
GEF\cite{wang_2020b} & 35 & 60 & \textbf{2480 $\times$ 2048} & \greencheck & 120 & 240 $\times$ 180 & \greencheck & \greencheck \\
\textbf{HS-ERGB~(Ours)}& \textbf{226} & \textbf{71.45} & 1440 $\times$ 1080 & \greencheck & 120 & \textbf{720 $\times$ 1280} & \greencheck & \greencheck \\
\hline
\end{tabular}}
\end{table*}%
\begin{figure*}[tb]
    \centering
    \begin{tabular}{ccc}
    \includegraphics[width=0.32\linewidth]{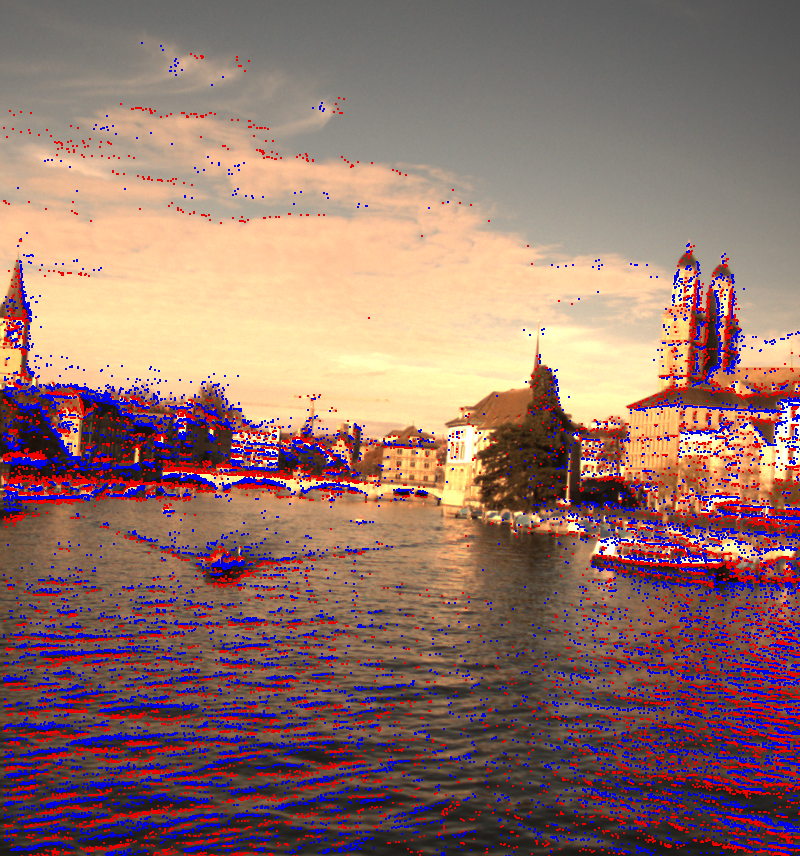}&
    \includegraphics[width=0.32\linewidth]{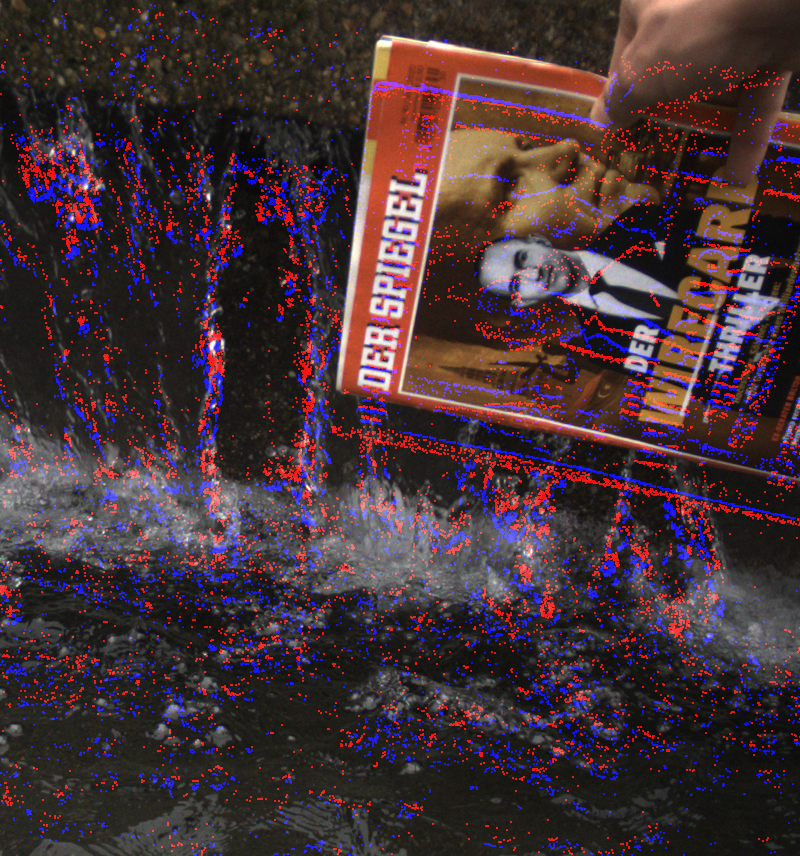}&
    \includegraphics[width=0.32\linewidth]{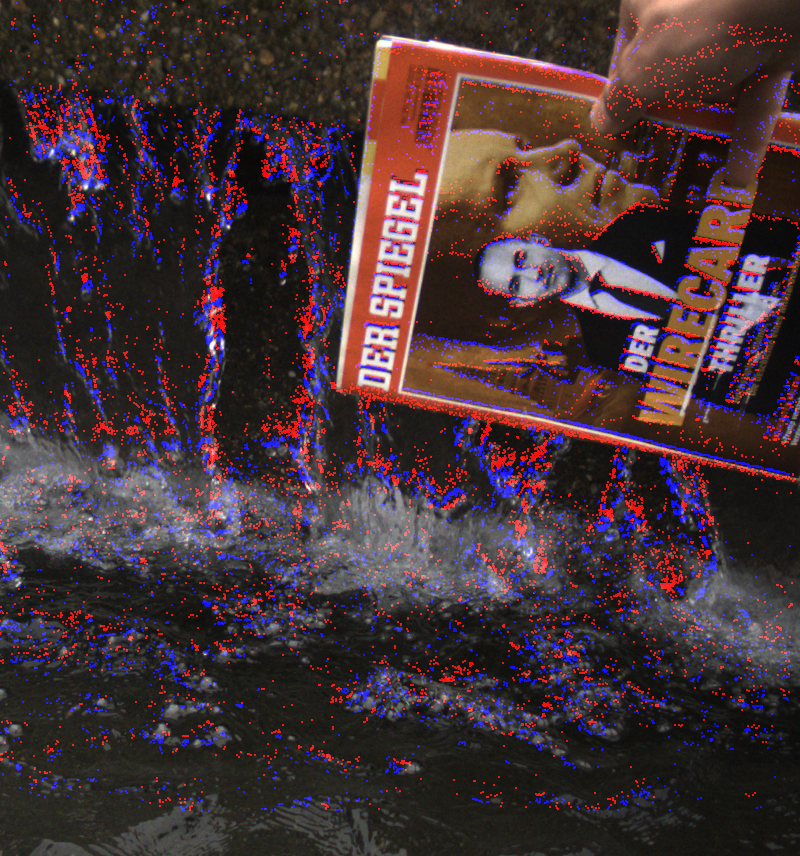}\\
    (a) far away scenes& (b) misaligned close scenes &(c) after global alignment
    \end{tabular}
    \caption{Alignment of standard frames with events. Aggregated events (blue positive, red negative) are overlain with the standard frame.   
    For scenes with sufficient depth (more than \SI{40}{\meter}) stereo rectification of both outputs yields accurate per-pixel alignment (a).
    However, for close scenes (b) events and frames are misaligned. In the absence of camera motion and motion in a plane, the views can be aligned with a global homography (c).}
    \label{fig:camera_alignment}    
\end{figure*}

To makes sure that the fine-tuning does not affect our general conclusions, we fine-tuned the proposed method and RRIN method~\cite{li_2020} on subset of High Quality Frames dataset and test them on the remaining part~(``poster\_pillar\_1'', ``slow\_and\_fast\_desk'', ``bike\_bay\_hdr'' and ``desk'' sequences).
We choose RRIN method for this experiment, because it showed good performance across synthetic and real datasets and it is fairly simple. As clear from the Tab.~\ref{tab:finetuning}, after the fine-tuning,  performance of the proposed method remained very strong compared to the RRIN method.

\section{High Speed Events and RGB Dataset}

\begin{table*}[tb]
   \caption{Overview of all sequences of the High  Speed Event-RGB~(HS-ERGB) dataset.}
    \label{tab:HS-ERGB_overview}
    \resizebox{1\textwidth}{!}{%
    \begin{tabular}{lcll}
    \hline
    \textbf{Sequence Name} & \textbf{Subset} & \textbf{Camera Settings} &  \textbf{Description} \\ 
    \hline
    \multicolumn{4}{l}{\textbf{Close planar sequences}} \\
    \hline
    Water bomb air (Fig.~\ref{fig:waterbombair})
    &  \multirow{5}{*}{Train} & 163 FPS, \SI{1080}{\micro\second} exposure, 1065 frames & accelerating object, water splash    \\ 
    \mbox{Lighting~match} 
    &        & 150 FPS, \SI{2972}{\micro\second} exposure, 666 frames &  illumination change, fire  \\
      
    \mbox{Fountain~Schaffhauserplatz~1} 
    &        & 150 FPS, \SI{977}{\micro\second} exposure, 1038 frames &  illumination change, fire  \\ 
    
    Water~bomb~ETH~2 (Fig.~\ref{fig:waterbombeth})
    &         & 163 FPS, \SI{323}{\micro\second} exposure, 3494 frames     &  accelerating object, water splash     \\  
    Waving~arms 
    &         & 163 FPS, \SI{3476}{\micro\second} exposure, 762 frames     &   non-linear motion  \\ 
    \hline
    Popping~air~balloon 
    &  \multirow{5}{*}{Test}       & 150 FPS, \SI{2972}{\micro\second} exposure, 335 frames &   non-linear motion, object disappearance \\
    Confetti~(Fig.~\ref{fig:confetti} 
    &         & 150 FPS, \SI{2972}{\micro\second} exposure, 832 frames & non-linear motion, periodic motion     \\  
    Spinning~plate 
    &         & 150 FPS, \SI{2971}{\micro\second} exposure, 1789 frames     &   non-linear motion, periodic motion    \\
    Spinning~umbrella 
    &         & 163 FPS, \SI{3479}{\micro\second} exposure, 763 frames   & non-linear motion    \\
    Water~bomb~floor~1~(Fig.~\ref{fig:waterbombfloor}) 
    &         & 160 FPS, \SI{628}{\micro\second} exposure, 686 frames   & accelerating object, water splash   \\  
    Fountain~Schaffhauserplatz~2 
    &         & 150 FPS, \SI{977}{\micro\second} exposure, 1205 frames   & non-linear motion, water  \\
    Fountain~Bellevue~2~(Fig.~\ref{fig:fountainbellevue}) 
    &         & 160 FPS, \SI{480}{\micro\second} exposure, 1329 frames   & non-linear motion, water, periodic movement \\ 
    Water~bomb~ETH~1 
    &         & 163 FPS, \SI{323}{\micro\second} exposure, 3700 frames   & accelerating object, water splash    \\
    Candle~(Fig.~\ref{fig:candle}) 
     &         & 160 FPS, \SI{478}{\micro\second} exposure, 804 frames   & illumination change, non-linear motion    \\ 
     \hline
    \multicolumn{4}{l}{\textbf{Far-away sequences}} \\
    \hline
        Kornhausbruecke~letten~x~1 &  \multirow{19}{*}{Train}  & 163 FPS, \SI{266}{\micro\second} exposure, 831 frames        & fast camera rotation around z-axis    \\ 
    Kornhausbruecke~rot~x~5    &        & 163 FPS, \SI{266}{\micro\second} exposure, 834 frames     &  fast camera rotation around x-axis     \\  
    Kornhausbruecke~rot~x~6    &        & 163 FPS, \SI{266}{\micro\second} exposure, 834 frames     &  fast camera rotation around x-axis     \\  
    Kornhausbruecke~rot~y~3   &         & 163 FPS, \SI{266}{\micro\second} exposure, 833 frames     &  fast camera rotation around y-axis     \\  
    Kornhausbruecke~rot~y~4   &         & 163 FPS, \SI{266}{\micro\second} exposure, 833 frames     &   fast camera rotation around y-axis     \\  
    Kornhausbruecke~rot~z~1   &         & 163 FPS, \SI{266}{\micro\second} exposure, 857 frames     &   fast camera rotation around z-axis     \\  
    Kornhausbruecke~rot~z~2   &         & 163 FPS, \SI{266}{\micro\second} exposure, 833 frames & fast camera rotation around z-axis     \\  
    Sihl~4   &         & 163 FPS, \SI{426}{\micro\second} exposure, 833 frames     &   fast camera rotation around z-axis    \\  
    Tree~3   &         & 163 FPS, \SI{978}{\micro\second} exposure, 832 frames   & camera rotation around z-axis    \\  
    Lake~4   &         & 163 FPS, \SI{334}{\micro\second} exposure, 833 frames   & camera rotation around z-axis   \\  
    Lake~5   &         & 163 FPS, \SI{275}{\micro\second} exposure, 833 frames   & camera rotation around z-axis   \\  
    Lake~7   &         & 163 FPS, \SI{274}{\micro\second} exposure, 833 frames   & camera rotation around z-axis    \\  
    Lake~8  &         & 163 FPS, \SI{274}{\micro\second} exposure, 832 frames   & camera rotation around z-axis  \\  
    Lake~9  &         & 163 FPS, \SI{274}{\micro\second} exposure, 832 frames   & camera rotation around z-axis      \\  
    Bridge~lake~4  &         & 163 FPS, \SI{236}{\micro\second} exposure, 836 frames   & camera rotation around z-axis      \\  
    Bridge~lake~5  &         & 163 FPS, \SI{236}{\micro\second} exposure, 834 frames   & camera rotation around z-axis      \\  
    Bridge~lake~6  &         & 163 FPS, \SI{235}{\micro\second} exposure, 832 frames   & camera rotation around z-axis      \\  
    Bridge~lake~7  &         & 163 FPS, \SI{235}{\micro\second} exposure, 832 frames   & camera rotation around z-axis      \\  
    Bridge~lake~8  &         & 163 FPS, \SI{235}{\micro\second} exposure, 834 frames   & camera rotation around z-axis      \\  
    \hline
    Kornhausbruecke~letten~random~4 &  \multirow{6}{*}{Test}       & 163 FPS, \SI{266}{\micro\second} exposure, 834 frames   & random camera movement      \\  
    Sihl~03   &       & 163 FPS, \SI{426}{\micro\second} exposure, 834 frames   & camera rotation around z-axis      \\  
    Lake~01   &         & 163 FPS, \SI{335}{\micro\second} exposure, 784 frames   & camera rotation around z-axis      \\  
    Lake~03   &         & 163 FPS, \SI{334}{\micro\second} exposure, 833 frames   & camera rotation around z-axis      \\  
    Bridge~lake~1   &         & 163 FPS, \SI{237}{\micro\second} exposure, 833 frames   & camera rotation around z-axis      \\  
    Bridge~lake~3   &         & 163 FPS, \SI{236}{\micro\second} exposure, 834 frames   & camera rotation around z-axis      \\  
    \hline
    \end{tabular}}
 
\end{table*}

\begin{figure*}[htbp]
     \centering
     \begin{subfigure}[b]{0.49\textwidth}
         \centering
         \animategraphics[width=0.9\linewidth,autoplay,loop]{5}{figures/dual_dataset_example/water_bomb_air_0000}{0}{14}
         \caption{Water bomb air}
         \label{fig:waterbombair}
     \end{subfigure}
     \begin{subfigure}[b]{0.49\textwidth}
         \centering
         \animategraphics[width=0.9\linewidth,autoplay,loop]{5}{figures/dual_dataset_example/fountain_bellevue2_0000}{0}{14}
         \caption{Fountain Bellevue}
         \label{fig:fountainbellevue}
     \end{subfigure}
     \begin{subfigure}[b]{0.49\textwidth}
         \centering
         \animategraphics[width=0.9\linewidth,autoplay,loop]{5}{figures/dual_dataset_example/water_bomb_eth_02_0000}{0}{14}
         \caption{Water bomb ETH 2}
         \label{fig:waterbombeth}
     \end{subfigure}
     \begin{subfigure}[b]{0.49\textwidth}
         \centering
         \animategraphics[width=0.9\linewidth,autoplay,loop]{5}{figures/dual_dataset_example/water_bomb_floor_01_0000}{0}{14}
         \caption{Water bomb floor 1}
         \label{fig:waterbombfloor}
     \end{subfigure}
     \begin{subfigure}[b]{0.49\textwidth}
         \centering
         \animategraphics[width=0.9\linewidth,autoplay,loop]{5}{figures/dual_dataset_example/confetti_}{0}{14}
         \caption{Confetti}
         \label{fig:confetti}
     \end{subfigure}
     \begin{subfigure}[b]{0.49\textwidth}
         \centering
         \animategraphics[width=0.9\linewidth,autoplay,loop]{5}{figures/dual_dataset_example/candle_0000}{0}{14}
         \caption{Candle}
         \label{fig:candle}
     \end{subfigure}
     \caption{Example sequences of the HS-ERGB dataset. This figure contains animation that can be viewed in Acrobat Reader.}\label{fig:dataset_animation}
 \end{figure*}

In this section we describe the sequences in the High-Speed Event and RGB~(HS-ERGB) dataset. The commercially available DAVIS 346 \cite{brandli_2014} already allows the simultaneous recording of events and grayscale frames, which are temporally and spatially synchronized. However, it has some shortcomings as the relatively low resolution of only $346 \times 260$ pixels of both frames and events. This is far below the resolution of typical frame based consumer cameras. Additionally, the DAVIS 346 has a very limited dynamic range of 55~db and a maximum frame of 40 FPS. Those properties render it not ideal for many event based methods, which aim to outperform traditional frame based cameras in certain applications. The setup described in \cite{wang_2020b} shows improvements in the resolution of frames and dynamic range, but has a reduced event resolution instead. 
The lack of publicly available high resolution event and color frame datasets and of the shelf hardware motivated the development of our dual camera setup. It features high resolution, high frame rate, high dynamic range color frames combined with high resolution events. A comparison of our setup with the DAVIS346\cite{brandli_2014} and the setup with beam splitter in \cite{wang_2020b} is shown in~\ref{tab:dual_setup_specs}. With this new setup we collect new High Speed Events and RGB~(HS-ERGB) Dataset that we summarize in Tab.~\ref{tab:HS-ERGB_overview}. We show several fragments from the dataset in Fig.~\ref{fig:dataset_animation}. In the following paragraphs we describe temporal synchronization and spatial alignment of frame and event data that we performed for our dataset.   

\textbf{Synchronization}
In our setup, two cameras are hardware synchronized through the use of external triggers. 
Each time the standard camera starts and ends exposure, a trigger is sent to the event camera which records an \emph{external trigger event}
with precise timestamp information. 
This information allows us to assign accurate timestamps to the standard frames, as well as group events during exposure or between consecutive frames.

\textbf{Alignment}
In our setup event and RGB cameras are arranged in stereo configuration, therefore event and frame data in addition to temporal, require spatial alignment. We perform the alignment in three steps: \emph{(i)} stereo calibration, \emph{(ii)} rectification and \emph{(iii)} feature-based global alignment. 
We first calibrate the cameras using a standard checkerboard pattern. 
The recorded asynchronous events are converted to temporally aligned video reconstructions using E2VID\cite{rebecq_2019a,rebecq_2019b}. 
Finally, we find the intrinsic and extrinsics by applying the stereo calibration tool Kalibr\cite{oth_2013} to the video reconstructions and the standard frames recorded by the color camera. 
We then use the found intrinsics and extrinsics to rectify the events and frames.

Due to the small baseline and similar fields of view (FoV), stereo rectification is usually sufficient to align the output of both sensors for scenes with a large average depth ($>$\SI{40}{\meter}).
This is illustrated in Fig. \ref{fig:camera_alignment} (a). 

For close scenes, however, events and frames are misaligned (Fig. \ref{fig:camera_alignment} (b)).
For this reason we perform the second step of global alignment using a homography which we estimate by matching SIFT features \cite{Lowe04ijcv} extracted on the standard frames and video reconstructions. The homography estimation also utilizes RANSAC to eliminate false matches.
When the cameras are static, and the objects of interest move within a plane, this yields accurate alignment between the two sensors (Fig.\ref{fig:camera_alignment} (c)).

{\small
\bibliographystyle{ieee_fullname}
\bibliography{main}
}

\end{document}